%% file: main.tex
\documentclass[10pt,twocolumn,letterpaper]{article}

\usepackage{iccv}      


\definecolor{iccvblue}{rgb}{0.21,0.49,0.74}
\usepackage[pagebackref,breaklinks,colorlinks,allcolors=iccvblue]{hyperref}

\def\confName{ICCV}
\def\confYear{2025}
\usepackage{indentfirst}
\title{Semantic Discrepancy-aware Detector for Image Forgery Identification}

\author{
Ziye Wang$^1$, Minghang Yu$^1$, Chunyan Xu$^{1\textsubscript{*}}$, Zhen Cui$^{2\textsubscript{*}}$\thanks{\hspace{-0.6cm}\textsuperscript{*}Corresponding Authors: Z. Cui and C. Xu. \\
Email: \{wzynjust,mhyu,cyx\}@njust.edu.cn (Z. Wang, M. Yu and C. Xu), zhen.cui@bnu.edu.cn (Z. Cui).}\\
$^1$Nanjing University of Science and Technology, Nanjing, Jiangsu, China\\
$^2$Beijing Normal University, Beijing, China
}

\makeatletter
\def\thanks#1{\protected@xdef\@thanks{\@thanks
        \protect\footnotetext{#1}}}
\makeatother

\usepackage{multirow} 

\begin{document}
\maketitle
\input{sec/0_abstract}    
\input{sec/1_intro}

\input{sec/2_formatting}

{
    \small
    \bibliographystyle{ieeenat_fullname}
    \bibliography{main}
}
\input{sec/X_suppl}

\end{document}

%% file: sec/0_abstract.tex
\begin{abstract}
With the rapid advancement of image generation techniques, robust forgery detection has become increasingly imperative to ensure the trustworthiness of digital media. Recent research indicates that the learned semantic concepts of pre-trained models are critical for identifying fake images. However, the misalignment between the forgery and semantic concept spaces hinders the model's forgery detection performance. To address this problem, we propose a novel \textbf{S}emantic \textbf{D}iscrepancy-aware \textbf{D}etector (SDD) that leverages reconstruction learning to align the two spaces at a fine-grained visual level. 
By exploiting the conceptual knowledge embedded in the pre-trained vision-language model, we specifically design a semantic token sampling module to mitigate the space shifts caused by features irrelevant to both forgery traces and semantic concepts. 
A concept-level forgery discrepancy learning module, built upon a visual reconstruction paradigm, is proposed to strengthen the interaction between visual semantic concepts and forgery traces, effectively capturing discrepancies under the concepts' guidance. 
Finally, the low-level forgery feature enhancemer integrates the learned concept-level forgery discrepancies to minimize redundant forgery information. Experiments conducted on two standard image forgery datasets demonstrate the efficacy of the proposed SDD, which achieves superior results compared to existing methods. The code is available at \url{https://github.com/wzy1111111/SSD}.

\end{abstract}

%% file: sec/1_intro.tex
\section{Introduction}
\label{sec:intro}

With the thriving of generative AI technologies, like Generative Adversarial Networks (GANs) \cite{GAN} and diffusion models \cite{DDIM}, the images generated by these methods can easily create confusion by passing off the spurious as genuine. Therefore, it is crucial to develop a universal method for detecting fake images to mitigate the widespread dissemination of disinformation.

\begin{figure}[!t]
    \centering
    \includegraphics[width=0.9\linewidth]{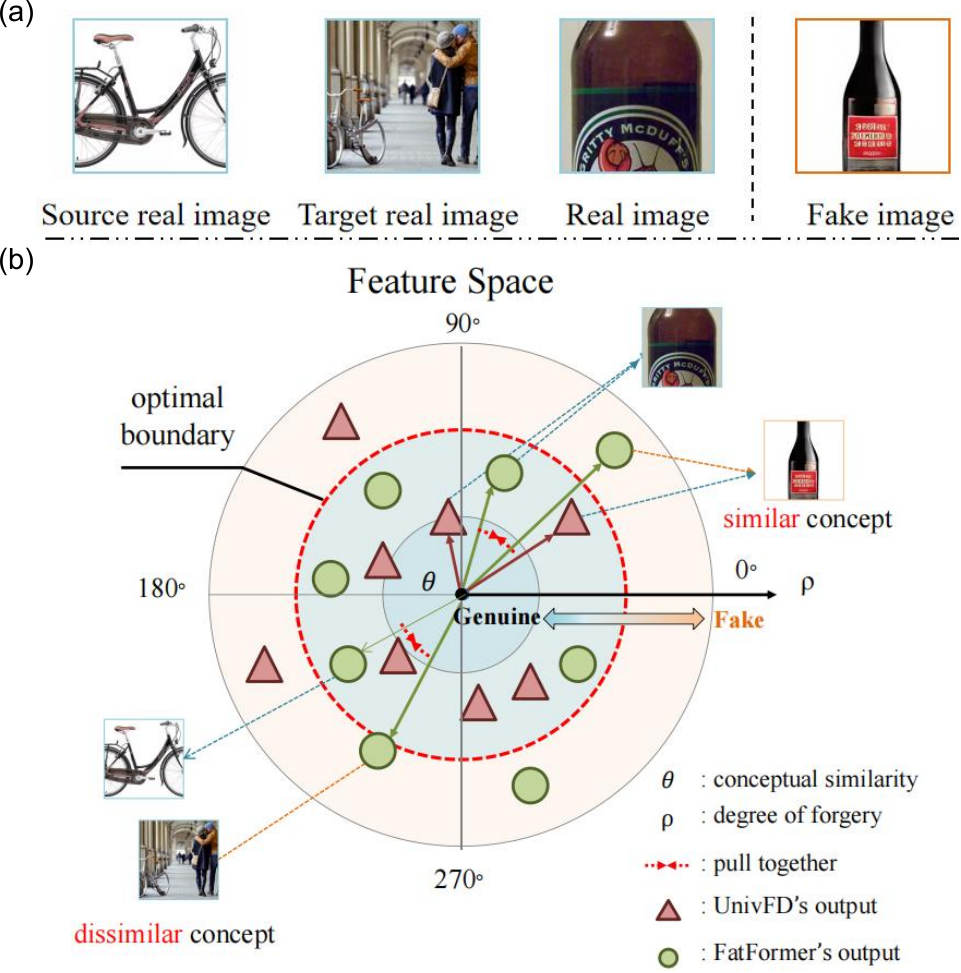}
    \caption{\textbf{The phenomenon of misalignment between semantic concept space and forgery space.} Since $\cos\theta$ can reflect the similarity of image descriptions, we model the feature space in polar coordinates. As the semantic concept space in \cite{UniFD} is frozen, fake samples sharing similar concepts with real ones can be easily misclassified. In the forgery-adaptive space like \cite{FatFormer},  the model can correctly distinguish between them based on re-learned forgery features. Nevertheless, due to the semantic concept bias introduced by coarse text prompts, the target samples may be projected into an inaccurate semantic concept dimension, causing them to drift away from the real source samples along the fake dimension.
    }\vspace{-0.6cm}
    \label{Reason}
\end{figure}

Pioneering research \cite{FatFormer, UniFD} has shown that projecting images into a joint embedding space of texts and images can effectively capture discrepancies between fake and real images. In contrast, previous methods \cite{grandient,patchfor,FreDect,CNNSpot} overlooking the interplay between forgery traces and semantic concepts perform poorly when confronted with unseen generative models.
To investigate the visual semantic concepts of pre-trained models, we conduct a statistical analysis of the output features from CNNSpot \cite{CNNSpot} and CLIP: ViT-L/4 \cite{UniFD} (See ~\cref{A} for more details). Under different categories, CNNSpot exhibits a synchronized difference between real and fake features in its feature space. However, when transitioning to the CLIP's space, these differences become inconsistent. From this, we infer a nuanced relationship between semantic concepts and forgery traces: different semantic concepts may guide the model to uncover distinct forgery traces.

Intuitively, relying on a frozen pre-trained vision-language model like UnivFD \cite{UniFD} is essential to  incorporate high-level semantic priors, but this tends to overlook fine-grained forgery details. Although FatFormer \cite{FatFormer} achieves a substantial enhancement in generalization by employing the forgery-aware adaptive transformer, we observe that soft prompts based on simple [CLASS] embeddings have an intrinsic limitation in their semantic description granularity (See ~\cref{B} for more details). The constrained breadth of the conveyed concepts may lead the detection toward incorrect predictions. This limitation highlights a misalignment between the visual semantic concept space and the target forgery space, as illustrated in ~\cref{Reason}. 
To address this issue, one empirical approach is to design more detailed text descriptions, but this method struggles to describe all visual details due to the limited length of texts and brings more computational overhead. Drawing from the aforementioned findings and analysis, we make a first attempt to align the CLIP’s visual semantic concept space with the forgery space by reconstructing semantic features.

\begin{figure}[tp]
    \centering
    \includegraphics[width=0.9\linewidth]{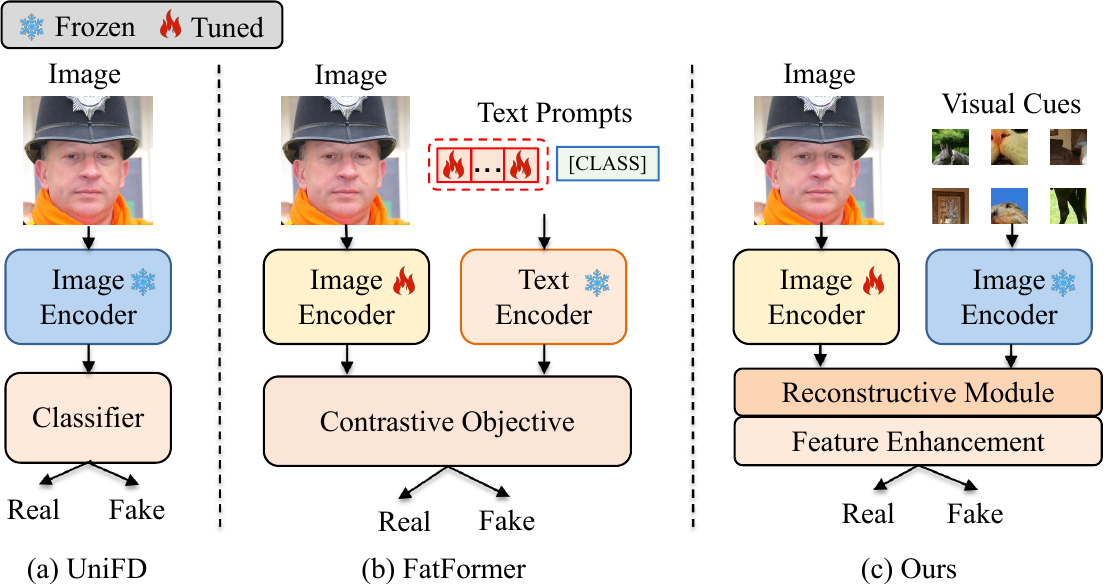}
    \caption{
    \textbf{Different paradigms of image forgery identification with pre-trained vision-language model.} (a)  Fine-tune the frozen model only by fully connected (FC) layers \cite{UniFD}. (b) Prompt-based designs are tuned on text prompts and contrastive objectives  \cite{FatFormer}. (c) Our paradigm incorporating visual clues can capture fine-grained forgery traces by reconstructive learning.} 
    \label{Paradigm}
    \vspace{-0.5cm}
\end{figure}

We develop a vision-based paradigm, as outlined in ~\cref{Paradigm}. First, employing a pre-trained model only with nearest neighbor or linear probing (\eg UnivFD \cite{UniFD}, ~\cref{Paradigm} (a)) is suboptimal for image forgery detection. Second, modifying the pre-trained model with task-specific prompts (\eg FatFormer \cite{FatFormer} in~\cref{Paradigm} (b)) may favor models biased towards any particular semantic concept. These studies pave the way for exploring the semantic concepts space. Inspired by image reconstruction \cite{Zamir2022LearningEF,Shen2022TransCSAT}, our paradigm amplifies the concept-level forgery discrepancies of forgery images, which empowers the model to detect suspicious forgery traces with the assistance of semantic concepts.

In this work, we present a novel Semantic Discrepancy-aware Detector (SDD) to accurately align the semantic concept space and the forgery space. 
To mitigate interference from features unrelated to both semantic concepts and forgery traces, we divide the real images into non-overlapping blocks and feed them to the frozen CLIP \cite{Radford2021LearningTV} to obtain diverse semantic patch tokens. These tokens acting as visual cues smoothly align the two spaces. It is noteworthy that these tokens sampled by JS divergence are universally representative of the real semantic distribution.
Then, the visual cues are fused into a concept-level forgery discrepancy module. Unlike FatFormer, LoRA layers are incorporated into the image encoder. The goal is to preserve the completeness and diversity of the learned semantic concepts of CLIP, while the forgery features sharing similar semantic concepts should be highlighted. During reconstruction, we only narrow the reconstruction gap for real samples to reinforce the reconstructed discrepancies of the synthetic images.
Finally, we introduce the low-level forgery
feature enhancer that leverages the reconstruction difference map to facilitate the extraction of the highly generalizable forgery features, while incurring minimal additional parameters. 
The main challenge is how to capture forgery features with strong semantic concept correlation and features with high forgery relevance but weak semantic concept ties to ensure the model converges to powerful features. Motivated by this, we apply convolutional modules and adaptive weight parameters to avoid over-relying on semantic concepts.

We thoroughly evaluate the generalization performance of our model on the UnivFD dataset \cite{UniFD} and the SynRIS dataset \cite{FakeInversion}. Surprisingly, our method achieves superior performance by a $ap_m$ of $98.51\%$ and a $acc_m$ of $93.61\%$ on the UnivFD benchmark \cite{UniFD} and an average AUROC of $95.1\%$ on the SynRIS benchmark \cite{FakeInversion}. In summary, our contributions are as follows:
\begin{itemize}
    \item We propose a robust Semantic Discrepancy-aware Detector (SDD) for forgery detection, specifically designed to align the semantic concept space and forgery space in terms of visual information.
    \item We sample semantic tokens to mitigate the space shifts and align the two spaces through reconstruction learning. Additionally, we strengthen low-level forgery features to enhance the model's robustness.
    \item Our method achieves superior performance on two benchmarks, demonstrating its superior capability in comparison to existing approaches. 
\end{itemize}

\begin{figure*}[!t]
    \centering
    \includegraphics[width=1.0\linewidth ]{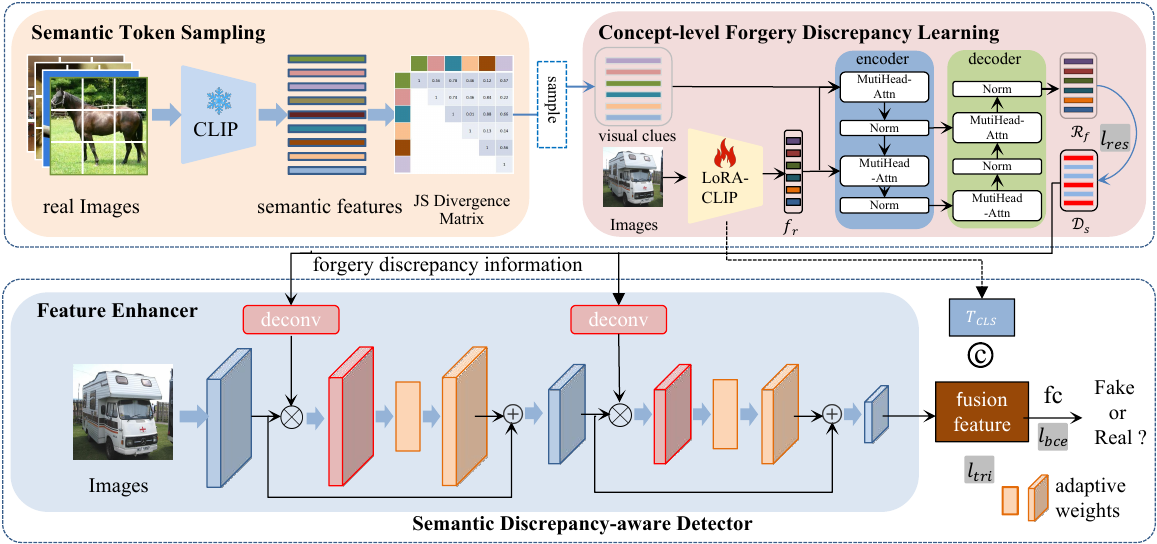}
    \caption{\textbf{The architecture of SDD. } We firstly sample semantic tokens from real images to learn features related to both semantic concepts and forgery. The input images are then mapped into a joint space of visual semantic concepts and forgery,  which are transformed into learnable features $V_H$.
    We use transformer-based encoder and decoder to get reconstructed features $\mathcal{R}_f$.
  A reconstruction difference map $\mathcal{D}_S$ is obtained and goes through the multi-scale convolutional network to refine forgery features. Finally, we concatenate the CLIP’s CLS token with this output along the same dimension for classification. The whole system is trained by jointly minimizing the binary cross-entropy loss $L_{bce}$, the reconstruction loss $L_{r}$, and the triplet loss $L_{tri}$. }\vspace{-0.4cm}
    \label{framework}
\end{figure*}

%% file: sec/2_formatting.tex
\section{Related Work}
\textbf{AI-generated images detection:} Extensive efforts have been devoted to enhancing the performance of AI-generated image detection. Early works like \cite{grandient,NPR,SpatialPhase} tend to mine the common forgery traces between all real and fake images, such as noise patterns, texture statistics, and frequency signals. As an illustration, Liu \etal \cite{Liu2022DetectingGI} designed a network that learns the consistent noise patterns in images for fake detection. Liu \etal\cite{texture} proposed to leverage the gram matrix to discover the global anomalous texture of fake images.  An effective approach \cite{FreDect} demonstrated that frequency representation is an important factor in improving fake detection performance. However, these differences are rigorously specific to the monotonous features, which contribute to the issue of overfitting. Cutting-edge research \cite{UniFD,FatFormer} shifted attention toward the semantic properties of images. Ojha \etal \cite{UniFD} showed that projecting images into the feature space of pre-trained vision-language model enables strong generalization ability. To build generalized forgery representations, Liu \etal \cite{FatFormer} constructed forgery adaptive space by a forgery-aware adapter. The above research \cite{UniFD,FatFormer,FakeInversion} has suggested that concept attributes are vital in the image forgery detection task. Assuming that diffusion-based models leave distinct forgery traces that are characteristic of specific concept distributions, we aim to extract robust forgery features guided by semantic concepts, rather than suppressing them. Therefore, even “useless” information can be useful by providing significant certainty about the content of the image. \\
\textbf{Reconstruction learning:} Reconstruction learning has great potential in unsupervised representation learning \cite{Han2019MultiAnglePC,Liu2020SPUNetSP}. Some works \cite{FakeInversion,AEROBL,re_1} utilized reconstruction learning to reveal the nuances between real and fake images. For example, Wang \etal \cite{Dire} found that reconstructing images by DDIM exposes an error between real images and their reconstructed replica. The new synthetic image detection method \cite{FakeInversion} used text-conditioned inversion maps to learn internal representations, which is conducive to predicting whether an image is fake. Ricker \etal \cite{AEROBL} offered a simple detection approach by applying autoencoders to measure the reconstruction error. Notably, these works are committed to mining the representation discrepancy of images by treating the representation space of generative models as a forgery reference distribution. Unlike previous works, we follow UnivFD to use a pretrained vision-language model as the real reference space to mine the feature divergence of images.

 \label{sec:intro}

\section{Methodology}
Our goal is to align the forgery and visual semantic concept spaces by reconstruction techniques for robust and generalizable synthetic image detection. To this end, we propose a fine-grained model named Semantic Discrepancy-aware Detector (SDD), which harness the generalization capability of vision-language models.

To better illustrate our method, we first introduce two key definitions involved in our framework. \textbf{Semantic concept space}: the ideal joint embedding space of images and texts with four properties: semantic alignment, modality invariance, locality consistency, and structure preservation. \textbf{Forgery space}: this denotes an ideal discriminative representation space that highlights forgery-specific traces. Notably, we derive the semantic concept space by a vision-language model pre-trained solely on real images and thus treat the two spaces as inherently independent.

In SSD, the Semantic Tokens Sampling (STS) module utilizes Jensen-Shannon (JS) divergence to sample semantic patch tokens, facilitating the model in accurately associating real and fake images. The Concept-level Forgery Discrepancy Learning (CFDL) module employs reconstruction learning to capture the forgery discrepancies within the fine-tuned semantic concept space, which focuses on identifying subtle variations in reconstructed forgery features. Finally, the reconstruction difference map is fed into Feature Enhancer module, which aims to refine low-level forgery features  with rich visual details.
  
\subsection{Semantic Tokens Sampling} 
Initially, we consider to directly align the visual semantic concept space and forgery space by leveraging fine-grained reconstruction learning to model model a joint distribution over semantic concepts and forgery traces. However, this strategy would treat the differences in features unrelated to semantics and forgery as crucial factors for identifying image's authenticity. To eliminate these redundant features, we sample real semantic image patch tokens as visual cues to bridge the semantic gap between real and fake images. This module enables the model to focus on concept-related forgery traces and highlight the distinctions between real and fake images.
In a tangible way, the image encoder of CLIP: ViT-L/14 is adapted to transform a real image $x_r$ into a set of features $f_{r}$, without the image CLS token. We define the transformation as:
\begin{equation}
	f_{r}=\phi(x_{r}),
\end{equation}
where $\phi(\cdot) $ refers to the CLIP:ViT-L/14's visual encoder, $x_{r} \in \mathbb{I}_{r}^{H \times W\times3}$ represents a real image characterized by a height of $H$ and a width of  $W$ . Besides, $f_{r} \in \mathbb{R}^{ N\times D}$, where $N$ is the number of tokens and  $D$ denotes the dimension of each patch token.

Since integrating all real patch tokens into the image reconstruction module is computationally intensive and memory-consuming, it is urgent to select a suitable subset of these tokens. From a distribution perspective, the Jensen–Shannon (JS) divergence, derived from the Kullback–Leibler divergence \cite{Erven2012RnyiDA}, is a symmetric and finite metric that can effectively measure the similarity between tokens by quantifying differences in their distributions. 
To calculate the JS divergence between two tokens, both are converted into computable probability distribution space using the softmax function. Let $f_{s} \in \mathbb{F}_{r}^{ M \times D}$ be the selected semantic patch tokens with the num of tokens $M=1/\delta$ and the dimension $D$ in terms of sampling rate $\delta$ ($0 \leq \delta \leq 1$, $\delta$ is user-defined). Once the initial token $\tilde{r}$ is determined, the JS divergence between $\tilde{r}$ and other tokens  falls within the range $\left[0,1\right]$. Subsequently, the sampling interval is split into $M$ equal segments with one token $r_j$ selected from each segment. As a consequence, the semantic tokens sampling module can be formulated as:
\begin{equation}
\begin{aligned}
     f_{s}&= \mathcal{S}(\mathbb{R}^{ N\times D},\delta)\\ 
     &=A_{c}^{N_{a} \times M} \times \mathbb{R}^{ N\times D} ,   \\
     s.t.\ 
     &\ a_{ij} = 1 \Rightarrow \text{JS}(\sigma(\tilde{r}),\sigma(r_j)) \in \left(\frac{j-1}{M}, \frac{j}{M}\right] , \\
     \  &\sum_{i=1}^{N_{a}}\sum_{j=1}^{M}a_{ij}=M , \sum_{i=1}^{N_{a}}a_{ij}\in \{0,1\} , \\
     &i=1,\dots,N_{a}, \ j=1,\dots,M,     
\end{aligned}
\end{equation}
where $S( \cdot, \cdot)$ represents the sampling process. $A_{c}^{N_{a} \times M}$ is a constraint matrix of size $N_{a} \times M$ whose element $a_{ij}$ is constrained to the binary pattern of $\left\{0, 1\right\}$. Here \text{JS} $(\cdot)$ refers to the Jensen-Shannon divergence, $N_{a}$ denotes the total number of real image patch tokens sampled from the training dataset of UnivFD and $M$ represents the required subset size. $\sigma(\cdot)$ refers to the softmax function. The sampling tokens help the reconstruction module avoid becoming biased towards any particular forgery-unrelated distribution. Meanwhile, it avoids the semantic bias often introduced by text prompts, since the tokens are evenly distributed in a unified CLIP space. 

\subsection{Concept-level Forgery Discrepancy Learning}
Coarse text prompts lack the semantic diversity, coverage, and detail necessary to convey the rich visual content of images, which in turn impairs the performance of image forgery detection. We argue that fine-grained visual details can uncover more forgery traces hidden in the images. As such, we mix sampling tokens with extracted features and capitalize on reconstruction learning to compensate for the omission of forgery traces brought by coarse prompts. As previous work \cite{FatFormer} has shown that the pre-trained vision-language model necessitates fine-tuning to adapt to the forgery detection task. Therefore, we integrate LoRA \cite{Hu2021LoRALA} with the CLIP-ViT model to capture discriminative forgery features by making use of the bread semantic concepts. This method, denoted as LoRA-CLIP \cite{Zanella2024LowRankFA}, is more streamlined and flexible. Given an input image $\mathcal{I} \in \mathbb{I}^{H \times W\times3}$, we  can get high-level visual features  $V_{H}$, as follows:
\begin{equation}
	V_{H}= \mathcal{F}_{LoRA}(\mathcal{I}).
\end{equation}
Here $\mathcal{F}_{LoRA}$ refers to the CLIP image encoder fine-tuned by LoRA. The reconstruction module of CFDL encompasses two submodules, i.e., transformer-based encoder and decoder. Thanks to the transformer's capability of long-range relationship modeling, we capitalize on the multi-head attention (MHA) mechanism, the core mechanism of transformer, to obtain more discriminative features by effectively incorporating contextual information. The MHA is set as:
\begin{equation}     
\left\{
\begin{aligned}
head_i=&\text{Attn}(QW_i^Q,KW_i^K,VW_i^V)\\
=&\sigma \left(\frac{QW_i^Q(KW_i^K)^\top}{\sqrt{d}}\right)VW_i^V,  \\
\text{MHA}(Q&,K,V)= \text{Concat}(head_1,\dots,head_h)W^O,
\end{aligned}
\right. 
\end{equation}
where $Q$(Query), $K$(Key), $V$(Value) refer to the input, $W_i^Q,  W_i^K, W_i^V$ separately denote the corresponding weights of linear projection, Attn($\cdot$) denotes the function of the scaled dot product, $d$ refers to the dimension of input, Concat ($\cdot$) represents the concatenation used to stitch the discrete attention outputs of head $1\sim h$ together.

 To amplify the discrepancy between a fake image and its reconstructed counterpart, the sampled visual clues are employed for the initial processing by the encoder. The encoding process can be formulated as follows:
\begin{align}
     R_1&=\text{LN}(\text{MHA}(f_{s},V_H,V_H)),\\
     R_2&=\text{LN}(\text{MHA}(R_1,V_H,V_H)),
\end{align}
where $\text{LN}(\cdot)$ denotes the Layer Normalization. Then, the encoder's outputs used as queries are injected into the decoder to get the final reconstructed features, which are similar to the encoder process and perform the following operation:
\begin{align}
     R_3&=\text{LN}(\text{MHA}(R_2,R_2,R_2)) ,\\
     R_{e}&=\text{LN}(\text{MHA}(R_1,R_3,R_3)).
\end{align}

During the reconstruction process, we just calculate the reconstruction loss $\mathcal{L}_r$ between the real input features and their reconstructed counterparts $\mathcal{R}_e$ within a mini-batch as follows:
\begin{equation}
	\mathcal{L}_{r}=\frac{1}{B}\sum_{i=0}^{B}\text{MSE}(R_{e},V_H),
\end{equation}
where $\text{MSE}(\cdot, \cdot)$ is the mean squared error. $\mathcal{L}_r$ encourages preserving the completeness and richness of the visual semantic concept space and highlighting the concept-related forgery features. 
Given the reconstructed features $R_{f}$ and the original feature $f_{r}$, the reconstruction difference map can be formally expressed as:
\begin{equation}
	\mathcal{D}_s=\vert R_{f}-f_{r}\vert,
\end{equation}
where $\vert\cdot$$\vert$ denotes the absolute value function. 

\subsection{Low-level Forgery Feature Enhancer}
Existing methods based on pre-trained vision-language models \cite{UniFD,FatFormer} overlook the importance of features with weak semantic relevance. We believe that a thorough alignment between the visual semantic concept space and the forgery space should include the exploration of forgery features that are weakly related to semantic concepts. Meanwhile, to eliminate redundant forgery features, we come up with a novel feature enhancer that refines low-level forgery features. Empowered by the reconstruction difference map, our detector orchestrates the extraction of multi-scale features with exceptional robustness and markedly enhanced effectiveness.
\begin{figure}[!t]
    \centering
    \includegraphics[width=5.5cm ]{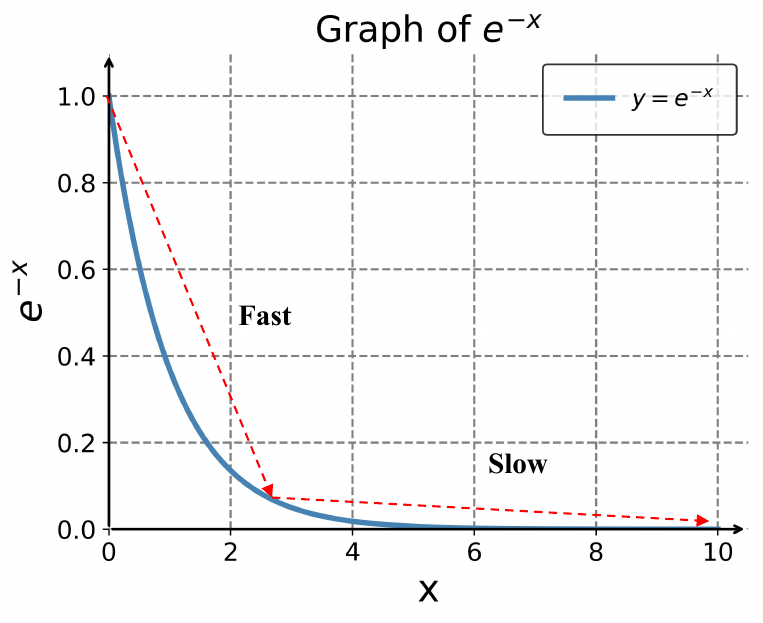}\vspace{-0.2cm}
    \caption{The curve of exponential inverse. In the “fast” interval, the value drops sharply. In the “slow” interval, the curve flattens out, indicating a decay toward 0. } 
    \label{efunction}
    \vspace{-0.5cm}
\end{figure}
As shown in ~\cref{framework}, the enhancer follows the typical architecture of a convolutional network. It involves the repeated application of convolutions, each followed by a batch normalization (BN) and a rectified linear unit (ReLU). For a given stage $n$, $F(n) $ ($n=1,2,3$) corresponds to its output features. We then deconvolve the semantic difference map $D_s$ to match the shape of $F(n)$ and perform pixel-wise multiplication with $F(n)$ to get $F^{'}(n)$ as:
\begin{equation}
	F^{'}(n) = F(n) \otimes \text{deconv}(D_s),
\end{equation}where $\otimes$ is the element-wise multiplication, $\text{deconv}(\cdot)$ represents deconvolution operation and $F^{'}(n)$ is the low-level feature aggregated with semantic information. To further enhance the reliability of the extracted features, we compute an adaptive weight coefficient $\frac{1}{e_n}$ to indicate the importance of $D_s$ to $F(n)$:
\begin{equation}
	\frac{1}{e_n}=\frac{1}{e^{\vert F^{'}(n)-F(n)\vert}}.
\end{equation}

\begin{table*}[!t] \vspace{-0.2cm}
    \setlength{\tabcolsep}{2pt}
    \renewcommand{\arraystretch}{1.4}
    \centering
    \resizebox{\textwidth}{24mm}{
    \begin{tabular}{l l c c c c c c c c c c c c c c c c c c c c c}
        \hline
        \multirow{3}{*}{Methods} & \multirow{3}{*}{Ref} & \multicolumn{6}{c}{GAN} 
        &\multirow{3}{*}{\shortstack[c]{Deep \\ [1ex] fakes}}& \multicolumn{2}{c}{Low level}& \multicolumn{2}{c}{Perceptual loss}& \multirow{3}{*}{Guided}& \multicolumn{3}{c}{LDM}& \multicolumn{3}{c}{Glide}& \multirow{3}{*}{Dalle} & \multirow{3}{*}{mAP} \\
        \cmidrule(r){3-8} \cmidrule(r){10-11} \cmidrule(r){12-13} \cmidrule(r){15-17} \cmidrule(r){18-20} 
         & & Pro- & Cycle-& Big-&Style- & Gau-& Star-&  &\multirow{2}{*}{SITD} & \multirow{2}{*}{SAN} &\multirow{2}{*}{CRN}  & \multirow{2}{*}{IMLE} &  & 200&200& 100& 100&50 & 100&  & \\

        & & GAN& GAN& GAN & GAN&GAN &GAN&  & &  &  &&  &Steps& w/cfg & Steps& 27 & 27& 10  & & \\
        
        \hline
        CNN-Spot&CVPR2020        &\textcolor{red}{100.0}&93.47&84.50&99.54&89.49&98.15&89.02&73.75&59.47&98.24&98.40&73.72&70.62&71.00&70.54&80.65&84.91&82.07&70.59&83.58\\ 
        PatchFor&ECCV2020        &80.88&72.84&71.66&85.75&65.99&69.25&76.55&76.19&76.34&74.52&68.52&75.03&87.10&86.72&86.40&85.37&83.73&78.38&75.67&77.73\\ 
        Co-occurrence&Elect.Imag.&\underline{99.74}&80.95&50.61&98.63&53.11&67.99&59.14&68.98&60.42&73.06&87.21&70.20&91.21&89.02&92.39&89.32&88.35&82.79&80.96&78.11\\ 
        Freq-spec&WIFS2019       &55.39&\textcolor{red}{100.0}&75.08&55.11&66.08&\textcolor{red}{100.0}&45.18&47.46&57.12&53.61&50.98&57.72&77.72&77.25&76.47&68.58&64.58&61.92&67.77&66.21\\ 
        Dire&ICCV2023            &\textcolor{red}{100.0}&83.59&81.50&96.50&81.70&99.88&95.73&62.51&69.98&97.31&98.62&79.53&75.52&73.42&76.45&86.28&89.00&88.34&51.35&83.54\\
        UnivFD&CVPR2023           &\textcolor{red}{100.0}&99.46&99.59&97.24&\underline{99.98}&99.60&82.45&61.32&79.02&96.72&99.00&87.77&99.14&92.15&99.17&94.74&95.34&94.57&97.15&93.38\\ 
        NPR&CVPR2024&\textcolor{red}{100.0}&99.50&96.50&\textcolor{red}{99.80}&96.80&\textcolor{red}{100.0}&92.20&73.10&78.70&87.20&64.80&65.80&{99.80}&\textcolor{red}{99.80}&{99.80}&\textcolor{red}{99.70}&\textcolor{red}{99.80}&\textcolor{red}{99.80}&98.60&92.19\\
        FatFormer & CVPR2024     &\textcolor{red}{100.0}&\textcolor{red}{100.0}&\textcolor{red}{99.98}&\underline{99.75}&\textcolor{red}{100.0}&\textcolor{red}{100.0}&\textcolor{red}{97.99}&\underline{97.94}&\underline{81.23}&\textcolor{red}{99.84}&\underline{99.93}&\underline{91.92}&\underline{99.83}&\underline{99.22}&\underline{99.89}&\underline{99.27}&\underline{99.50}&\underline{99.33}&\textcolor{red}{99.84}&\underline{98.18} &\\
       \textbf{Ours}&&\textcolor{red}{100.0}&\underline{99.77}&\underline{99.93}&{99.48}&\underline{99.98}&\underline{99.97}&\underline{97.23}&\textcolor{red}{97.91}&\textcolor{red}{93.10}&\underline{99.79}&\textcolor{red}{99.96}&\textcolor{red}{92.06}&\textcolor{red}{99.88}&{98.95}&\textcolor{red}{99.92}&98.06&98.29&97.73&\underline{99.81}&\textcolor{red}{98.51}\\ \hline
       \end{tabular}}
    \caption{Average precision comparisons on the UnivFD dataset. We replicate the results of CNNSpot, Patchfor, Co-occurrence, Freq-spec, and UnivFD from the work \cite{UniFD}. Additionally, the results of Dire and NPR are obtained from models retrained by ourselves, whereas those of FatFormer come from the official pre-trained model.  
    \textcolor{red}{Red} and \underline{underline} indicate the best and the second best result, respectively.}
    \label{ap}
\end{table*}

\begin{table*}[!t]

    \setlength{\tabcolsep}{2pt}
    \renewcommand{\arraystretch}{1.4}
    \centering
    \resizebox{\textwidth}{24mm}{
    \begin{tabular}{l l c c c c c c c c c c c c c c c c c c c c c}
        \hline
        \multirow{3}{*}{Methods} & \multirow{3}{*}{Ref} & \multicolumn{6}{c}{GAN} 
        &  \multirow{3}{*}{\shortstack[c]{Deep \\ [1ex] fakes}}& \multicolumn{2}{c}{Low level}& \multicolumn{2}{c}{Perceptual loss}& \multirow{3}{*}{Guided}& \multicolumn{3}{c}{LDM}& \multicolumn{3}{c}{Glide}& \multirow{3}{*}{Dalle} & \multirow{3}{*}{Avg-acc} \\
        \cmidrule(r){3-8} \cmidrule(r){10-11} \cmidrule(r){12-13} \cmidrule(r){15-17} \cmidrule(r){18-20} 
        
        & & Pro- & Cycle-& Big-&Style- & Gau-& Star-&  &\multirow{2}{*}{SITD} & \multirow{2}{*}{SAN} &\multirow{2}{*}{CRN}  & \multirow{2}{*}{IMLE} &  & 200&200& 100& 100&50 & 100&  & \\

        & & GAN& GAN& GAN & GAN&GAN &GAN&  & &  &  &&  &Steps& w/cfg & Steps& 27 & 27& 10  & & \\
        
        \hline
        CNN-Spot&CVPR2020        &\underline{99.99}&85.20&70.20&85.70&78.95&91.70&53.47&66.67&48.69&86.31&86.26&60.07&54.03&54.96&54.14&60.78&63.80&65.66&55.58&69.58\\ 
        PatchFor&ECCV2020        &75.03&68.97&68.47&79.16&64.23&63.94&75.54&75.14&75.28&72.33&55.30&67.41&76.50&76.10&75.77&74.81&73.28&68.52&67.91&71.24\\ 
        Co-occurrence&Elect.Imag.&97.70&63.15&53.75&92.50&51.10&54.70&57.10&63.06&55.85&65.65&65.80&60.50&70.70&70.55&71.00&70.25&69.60&69.90&67.55&66.86\\ 
        Freq-spec&WIFS2019       &49.90&99.90&50.50&49.90&50.30&99.70&50.10&50.00&48.00&50.60&50.10&50.90&50.40&50.40&50.30&51.70&51.40&50.40&50.00&55.45\\ 
        Dire&ICCV2023            &99.86&73.47&60.68&72.39&65.15&93.60&88.86&52.78&56.39&\underline{90.07}&\underline{94.05}&61.05&59.35&59.95&60.65&69.30&72,70&71.00&52.75&71.19\\
        UnivFD&CVPR2023           &\textcolor{red}{100.0}&\underline{98.50}&94.50&82.00&\textcolor{red}{99.50}&97.00&66.60&63.00&57.50&59.50&72.00&70.03&94.19&73.76&94.36&79.07&79.85&78.14&86.78&81.38\\ 
        NPR&CVPR2024             &99.80&92.00&89.50&96.30&87.60&\underline{99.70}&79.40&61.40&\underline{70.60}&74.50&57.10&55.23&97.40& \textcolor{red}{98.70}&97.90&\textcolor{red}{97.00}&\textcolor{red}{97.90}&\textcolor{red}{97.00}&88.80&86.20\\
        FatFormer &CVPR2024&99.89&\textcolor{red}{99.36}&\textcolor{red}{99.50}&\underline{97.12}&\underline{99.43}&\textcolor{red}{99.75}&\textcolor{red}{93.25}&\underline{81.39}&68.04&69.47&69.47&\underline{76.00}&\textcolor{red}{98.55}&\underline{94.85}&\textcolor{red}{98.60}&\underline{94.30}&\underline{94.60}&\underline{94.15}&\textcolor{red}{98.70}&\underline{90.86}\\
        \textbf{Ours}&&99.88&95.76&\underline{96.70}&\textcolor{red}{98.08}&98.46&99.17&\underline{91.82}&\textcolor{red}{83.61}&\textcolor{red}{77.45}&\textcolor{red}{95.40}&\textcolor{red}{96.47}&\textcolor{red}{79.55}&\underline{98.05}&{94.60}&\underline{98.25}&92.20&93.35&91.80&\underline{98.00}&\textcolor{red}{93.61}\\
        \hline
    \end{tabular}}
    \caption{Accuracy comparisons with different methods on the UnivFD dataset.}
    \vspace{-0.6cm}
    \label{acc}
\end{table*}

Here we explain the role of the exponential inverse through ~\cref{efunction}. As $x$ grows large, the curve of $e^{-x}$ becomes flatter. In the “fast” interval, forgery features with a significant divergence from the semantic difference map will be assigned smaller weights, which mobilizes the network to capture concept strongly-related features. However, in the “slow” interval, features strongly associated with forgery can avoid being misguided by semantic concepts, which indicates that the order of importance is reversed. Next, we obtain the attended low-level features $F_{low}$ by the residual connection:
\begin{equation}
	F_{low}(n)=F(n)+\frac{F^(n)}{e_n}.
\end{equation}
\label{sec:method}


\vspace{-0.3cm}
For optimizing the anchor features $f_{a}$ of the enhancer, the following triplet loss \cite{FcaNet}  is employed to bring positive samples $f_{p}$ closer while pushing negative samples $f_{n}$ apart:
\begin{equation}
	\mathcal{L}_{tri}=\max(0,d(f_{p},f_{a})-d(f_{n},f_{a})+\alpha),
\end{equation}
where $d (\cdot) $ represents the Euclidean distance between samples and $\alpha$ is the margin.
On top of that, we concatenate the LoRA-CLIP's CLS token $T_{CLS}$ with $F_{low}$ along the same dimension to yield the refined representation $F_{out}$. This ensures that forged features exhibit distinctiveness across different semantic identities while preserving their uniformity within similar semantic identities. The process is formulated as: 
\begin{equation}
	F_{out}=F_{low} ||  T_{CLS},
\end{equation}
where $F^{out}$  is strategically integrated with a linear classifier to enable the execution of binary classification.
Eventually, the total loss function $\mathcal{L}$ of the proposed framework can be defined as:
\begin{equation}
	\mathcal{L}=\mathcal{L}_{bce}+\lambda_1\mathcal{L}_{tri}+\lambda_2\mathcal{L}_{r},
\end{equation}
where $\mathcal{L}_{bce}$ and $\mathcal{L}_{tri}$  refer to the binary cross-entropy loss and the triplet loss. $\mathcal{L}_{tri}$ and $\mathcal{L}_{r}$ are scaled by the hyper-parameters $\lambda_1$ and $\lambda_2$, respectively.

\section{Experiments}
\subsection{Experiment Setups}
\textbf{Datasets:} We follow the protocol described in \cite{UniFD}, using ProGAN's images as training data. Additionally, we adopt the protocol from \cite{FakeInversion}, where the training data is composed of fake Stable Diffusion v1 images \cite{stable} and random real LAION images \cite{laion}. The UnivFD dataset \cite{UniFD} covers a broad range of generative models, including GANs and diffusion models, such as ProGAN \cite{ProGAN},  StyleGAN \cite{StyleGAN}, BigGAN \cite{biggan}, CycleGAN \cite{CycleGAN}, StarGAN \cite{stargan}, GauGAN \cite{gaugan}, CRN \cite{crn}, IMLE \cite{imle}, SAN \cite{san}, SITD \cite{sitd}, DeepFakes \cite{deepfake}, Guided \cite{guided}, Glide \cite{glide}, LDM \cite{ldm} and DALL-E \cite{dalle}. The SynRIS dataset \cite{FakeInversion} is designed to avoid bias toward any specific topic, theme, or style and contains high-fidelity images generated by text-to-image models, such as  Kandinsky2 \cite{k2}, Kandinsky3 \cite{k3}, PixArt-$\alpha$ \cite{pixart}, SDXL-DPO \cite{dpo}, SDXL \cite{sdxl},
SegMoE \cite{segmoe}, SSD-1B \cite{ssd}, Stable-Cascade \cite{wchen}, Segmind-Vega \cite{vega},
and \text{Würstchen}2 \cite{wchen}, Midjourney \cite{midjourney}, DALL.E 3 \cite{dalle3} and playground \cite{playground}. As standard evaluation metrics, the average precision (AP), the accuracy (ACC) and  AUCROC are considered to measure the effectiveness of different methods.

\textbf{Baselines:} We perform comparisons with state-of-the-art methods, as follows: 
1) CNNSpot \cite{CNNSpot} relies only on a CNN. 
2) PatchFor \cite{patchfor} performs detection on a patch level.
3) Co-occurrence \cite{cooocurance} converts input images into co-occurrence matrices for detection.
4) Freq-spec \cite{fredet} employs the frequency spectrum of images.
5) Dire \cite{Dire} exploits the error between input images and its reconstruction counterpart. 
6) UnivFD \cite{UniFD} uses the pre-trained language-vision model to determine the authenticity of images.
7) NPR \cite{NPR} captures the generalized artifacts according to the local interdependence among image pixels.
8) FatFormer \cite{FatFormer} aims at extracting forgery-adaptive features based on UnivFD.
9) Fakeinversion \cite{FakeInversion} employs text-conditioned inversion maps extracted from Stable Diffusion.

\begin{table}[tp]
\centering
\footnotesize
\renewcommand{\arraystretch}{1.3}
\setlength{\tabcolsep}{1pt}
 \begin{tabular}{lccccccccc}
        \toprule
        \multirow{2}{*}{Methods}&CNN-
&Freq-&\multirow{2}{*}{Dire}&Univ&\multirow{2}{*}{NPR}&FatF&FakeIn&Patch&\multirow{2}{*}{\textbf{Ours}}\\
       &Spot&spec&&FD&&ormer&version&For&\\
        \toprule
Kandinsky2 &60.0&57.0&71.6&56.2&\textcolor{red}{97.5}&75.6&69.9&53.5&\underline{97.1}\\

Kandinsky3&65.9&45.7&74.9&61.4&\underline{93.7}&80.1&74.3&51.4&\textcolor{red}{94.8}\\
    
PixArt-$\alpha$&62.7&56.4&81.5&64.7&\underline{89.5}&75.3&73.0&49.6&\textcolor{red}{90.2}\\

SDXL-DPO&84.3&69.8&69.9&70.2&\textcolor{red}{97.6}&86.0&88.1&54.5&\underline{95.5}\\
Segmind-&\multirow{2}{*}{74.2}&\multirow{2}{*}{65.3}&\multirow{2}{*}{81.0}&\multirow{2}{*}{62.3}&\textcolor{red}{\multirow{2}{*}{97.1}}&\multirow{2}{*}{82.4}&\multirow{2}{*}{81.1}&\multirow{2}{*}{53.1}&\multirow{2}{*}{\underline{97.0}}\\
        Vega&&&&&&&&&\\
        SDXL&81.4&61.2&86.2&66.3&\underline{96.0}&85.1&80.7&63.9&\textcolor{red}{97.6}\\
        Seg-MoE&66.3&54.6&71.9&62.0&\underline{93.6}&70.8&71.3&49.8&\textcolor{red}{97.6}\\
        SSD-1B&72.6&67.8&79.8&62.8&\underline{99.6}&70.1&79.4&61.2&\textcolor{red}{99.8}\\
        Stable-&\multirow{2}{*}{70.5}&\multirow{2}{*}{62.1}&\multirow{2}{*}{74.1}&\multirow{2}{*}{68.2}&\multirow{2}{*}{\underline{97.6}}&\multirow{2}{*}{81.6}&\multirow{2}{*}{74.9}&\multirow{2}{*}{57.4}&\textcolor{red}{\multirow{2}{*}{99.2}}\\
        Cascade&&&&&&&&&\\
    \text {Würstchen}2&61.0&63.3&74.2&69.7&\underline{90.9}&72.9&70.5&
        47.2&\textcolor{red}{98.2}\\
Midjourney&63.0&50.9&\underline{72.4}&59.2&58.5&73.6&66.4&53.7&\textcolor{red}{90.0}\\
Playground&58.2&52.3&67.9&58.7&\textcolor{red}{93.1}&81.4&62.5&54.1&\underline{92.8}\\
DALL·E3 &71.6&59.9&\underline{80.8}&48.0&69.1&79.2&75.9&50.1&\textcolor{red}{85.9}\\
Average&68.6&58.9&75.9&62.3&\underline{90.3}&78.0&74.5&53.8&\textcolor{red}{95.1}\\
      \toprule
    \end{tabular} 
    \caption{AUROC comparisons with different methods on the SynRIS dataset. We retrieve the results of CNNSpot, UnivFD, and FakeInversion from \cite{FakeInversion} and obtain the results for Dire, NPR, FatFormer, and PatchFor using re-implemented models.  \textcolor{red}{Red} and \underline{underline} indicate the best and the second best result, respectively.}
    \label{auc}
    \vspace{-0.4cm}
\end{table}

\textbf{Implement details:} 
Our training settings are adapted from the approach outlined in the previous study \cite{FatFormer} with several key modifications. Specifically, early stopping is employed during model training, with an initial learning rate of $1 \times 10^{-5}$ and a batch size of 32. Additionally, the Lora layers are configured with hyperparameters $lora_r=6$, $lora_\alpha=6$, and a dropout rate of 0.8, while $\alpha$ is set to 8.0. The proposed method is implemented using Pytorch on 2 Nvidia GeForce RTX A6000 GPUs. \\

\begin{table}[tp]
\centering
\footnotesize  
\begin{tabular}{c|c c c|c c}
\toprule 
\multirow{2}{*}{\#}&STS&CFDL&Feature&\multicolumn{2}{c}{UnivFD \phantom{a}Dataset}\\
&\phantom{a}module&\phantom{a}module&Enhancer&\phantom{a}$ap_{m}$&$acc_{m}$\\
\toprule
1& &\checkmark&& 97.37&81.64\\
2& &\checkmark& \checkmark&\underline{97.41}&\underline{90.17}\\
3& \checkmark& \checkmark& &97.39&89.98\\
4& \checkmark&\checkmark &\checkmark &\textcolor{red}{98.52}&\textcolor{red}{93.61}\\
\toprule
\end{tabular}
\caption{Ablation study of the proposed modules on the UnivFD Dataset. We show the mean accuracy ($acc_{m}$) and average precision ($ap_{m}$). Red and underline indicate the best and the second-best result,
respectively. }
\label{module}
\vspace{-0.6cm}
\end{table}

\vspace{-0.5cm}
\subsection{Comparision Results} 
\begin{figure*}[!t]
	\begin{minipage}[t]{0.3\linewidth}
		\centering
		\includegraphics[width=4cm,height=3.5cm]{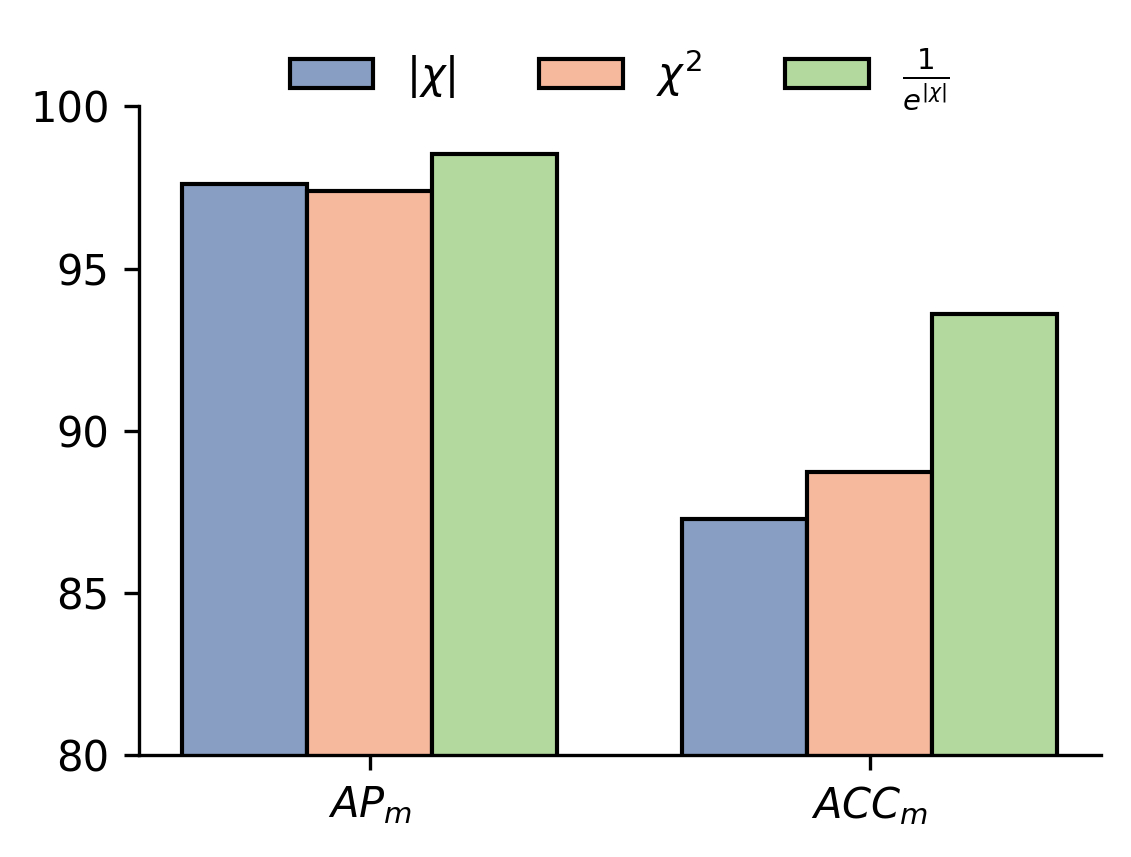}
            \caption{Performance of different functions on adaptive weights. }		    
            \label{ex}
	\end{minipage}\hspace{0.1cm}
	\begin{minipage}[t]{0.7\linewidth}
		\centering 
		\includegraphics[width=12cm,height=3.5cm]{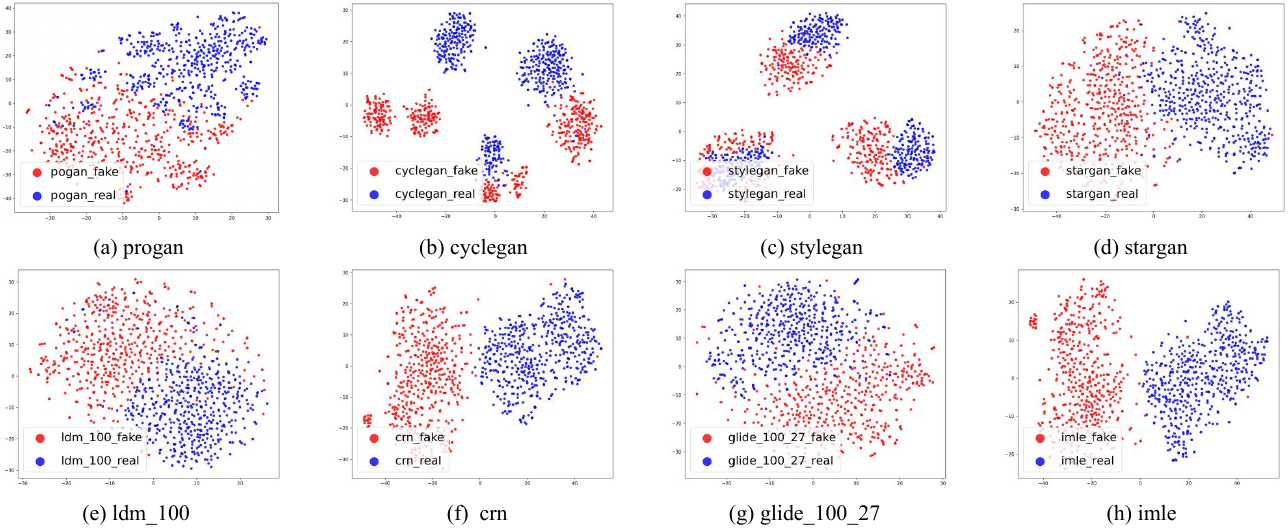}
             \caption{T-SNE visualization of real and fake images \cite{tsne}. The feature space is based on our classifier. Each randomly samples 500 real and 500 fake images.}
    \label{t-sne}
	\end{minipage}  
\end{figure*}

The UnivFD dataset includes a diverse range of models, allowing for a comprehensive evaluation of our method across both GAN and diffusion generative models. In addition, the SynRIS dataset provides images generated by cutting-edge generative models. 

\textbf{Results on the UnivFD dataset:} As reported in Table~\ref{ap} and Table~\ref{acc}, experimental results show that our proposed method achieves superior performance compared to exiting methods. Notably, without the biased interpretation introduced by coarse-grained text prompts, SDD surpasses the latest state-of-the-art method FatFormer by the mean AP ($ap_{m}$) of $0.34\%$ and the mean acc ($acc_{m}$) of $2.75\%$.  Compared with methods based on relatively monotonous forgery features, our approach can outperform all of them with a large improvement. The above evidence indicates effective combination of visual concepts and forgery features can contribute model to extract sufficient forgery traces and eliminate the superfluous features. 

\textbf{Results on the SynRIS dataset:} 
As shown in Table~\ref{auc}, when confronted with high-fidelity images generated by text-to-image models, methods leveraging pre-trained vision-language models, such as UnivFD and FatFormer, lose their competitiveness. In contrast, NPR, which focuses on neighboring pixel relationships, retains its edge. We assume that current generative methods grasp the relationships between visual information and semantic concepts in images but cannot refine local visual details at the pixel level. Considering that excessive reliance on concepts misses abnormal pixel arrangements and focusing on monotonous forgery patterns can cause overfitting, our detector, which emphasizes low-level features with visual concepts, is trained on lower-fidelity fake images generated by Stable Diffusion \cite{stable} to capture concept-specific lacunae. We follow the evaluation protocol from SynRIS \cite{FakeInversion}. In comparison, our detector achieves an impressive mean AUROC of $95.1\%$, surpassing the state-of-the-art method by $4.8\%$. This demonstrates its superior ability to tackle the challenges posed by evolving generative models.

\begin{figure}[tp]
    \centering
    \includegraphics[width=1\linewidth]{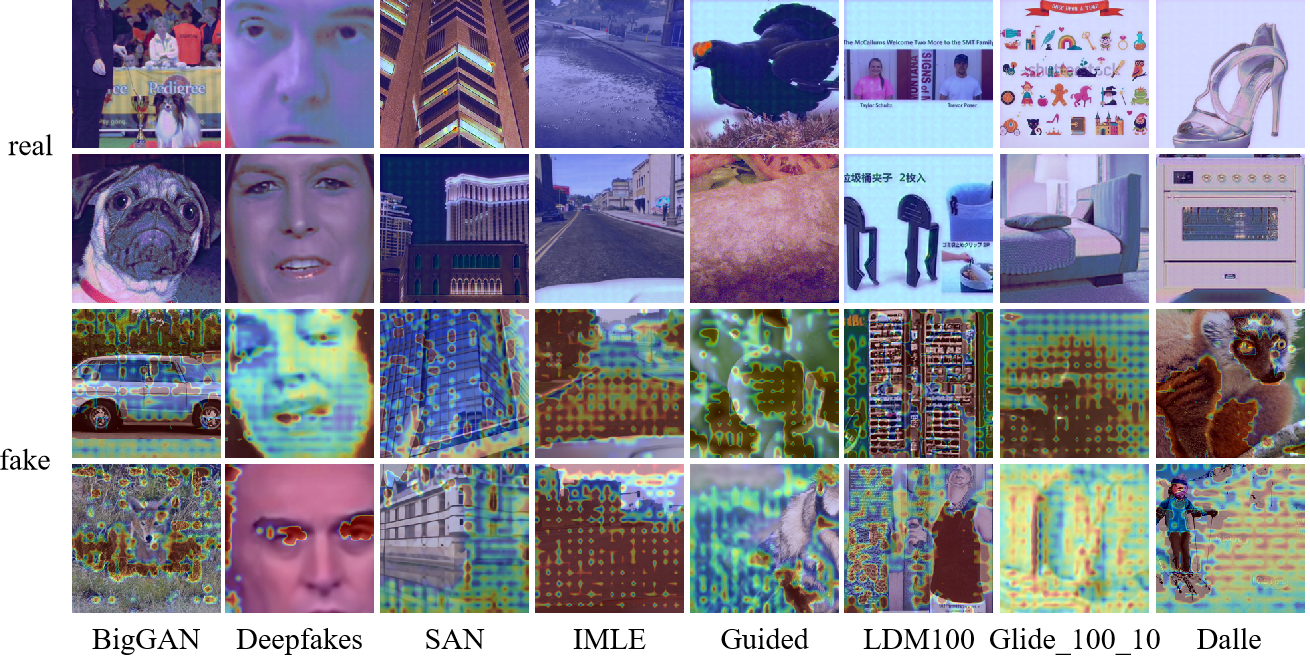}
    \caption{The showcase of attention maps for the input images.}
    \label{cam}\vspace{-0.2cm}
\end{figure}

\subsection{More analysis}
We perform comprehensive ablation studies on the UnivFD dataset under the original experimental configurations, reporting the mean accuracy ($acc_{m}$) and mean average precision ($ap_{m}$) as the primary evaluation metrics.


\textbf{Effect of each component:} 
We study the effects of removing STS module, CFDL module, and Feature Enhancer in our method. The results, presented in Table~\ref{module}, demonstrate that these components are essential for improving performance in generalization on unseen models. This empirical finding suggests that  CFDL effectively captures forgery discrepancies associated with semantic concepts, while the enhancer plays a crucial role in identifying robust forgery artifacts. The collaboration of all modules enhances the model's ability to distinguish between real and fake images.

\textbf{Effect of function on adaptive weights:}
To check how well the proposed function works in the SDD, we select two conventional functions for comparison: $f(x)=|x|$ and $f(x)=x^2$. The corresponding results are presented in Table~\ref{ex}. We find  our proposed function yields improvements in both $ap_{m}$ and $acc_{m}$ compared to the selected functions. These results demonstrate that our proposed function is effective in capturing robust and distinctive forgery features.


\textbf{Visualization of learned latent space:}  
As shown in ~\cref{t-sne}, the input images can be distinctly categorized into two clusters: real and fake. Nevertheless, why does the divided boundary of ProGAN appear ambiguous in contrast to other models? Additionally, why do the real clusters of CycleGAN and StyleGAN separate from each other? We attribute these to the influence of visual semantic concepts. Perceptively, with the supervision of visual semantic concepts, the learned boundary of ProGAN is more complex and nuanced, rather than just simple straight lines or curves. Similarly, the images generated by StyleGAN and CycleGAN are projected into the corresponding semantic concept distribution and then separated from the real images based on the visual semantic concepts.

\textbf{Visualization of attention on images:}          Fig.~\ref{cam} illustrates the attention maps for the input images, where we apply class activation mapping \cite{Zhou2015LearningDF} to visualize the learned representations. 
With the aid of semantic information, our model can focus on different regions of fake images, including the background, local object regions, and marginal details. This suggests that our fine-grained model is capable of capturing intricate discrepancies generalized to unseen models. Notably, the real images nearly always show no forgery discrepancy regions, which demonstrates the effectiveness of the reconstruction loss in the forgery detection task.

\vspace{-0.1cm}
\section{Conclusion}
In this paper, we propose a novel method, SDD, for generalizable forgery image detection. The findings show that our method establishes a new state-of-the-art in detecting images generated by generative models from different periods, which underscores its robustness and superior generalization capability. 
To the best of our knowledge, in pre-trained vision-language paradigms, our approach is the first to rely solely on visual information, without text prompts. Based on experimental results, we conclude that leveraging sampled tokens and reconstruction techniques effectively aligns the visual semantic concept space with the forgery space. Additionally, refining low-level forgery features under the supervision of visual semantic concepts enhances the performance of forgery detection. Although SDD performs well across various generative methods, there is still room for improvement as generative technologies continue to advance. 

\section*{Acknowledgement}
This work is supported by the National Natural Science
Foundation of China (Grants Nos. 62372238,62476133).

%% file: sec/X_suppl.tex
\clearpage
\setcounter{page}{1}
\maketitlesupplementary
\appendix
\section{Statistical analysis of feature values from detectors}
\label{A}
\begin{figure*}[htbp]
    \centering
    \includegraphics[width=\linewidth ]{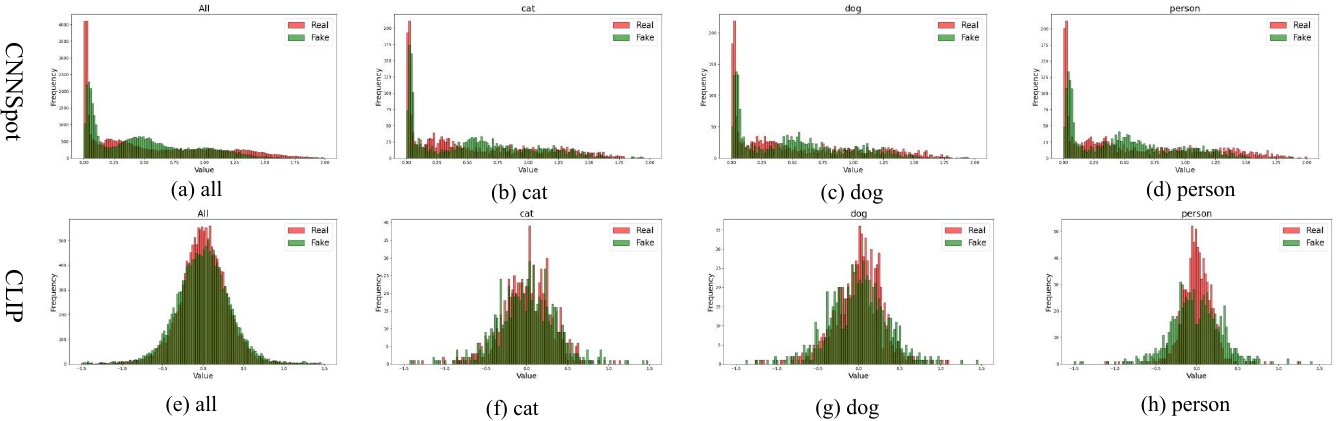}
    \caption{\textbf{Value statistics of extracted features.} We compare the input features from the last FC layer of CNNSpot \cite{CNNSpot} and CLIP \cite{UniFD}, both of which are fed with ProGAN \cite{AEROBL} data. Three classes from ProGAN's testing data are considered: cat, dog, and person. We also present the results for data from all classes. }
    \label{demograph}
\end{figure*}

\begin{figure*}[htbp]
    \centering
    \includegraphics[width=\linewidth ]{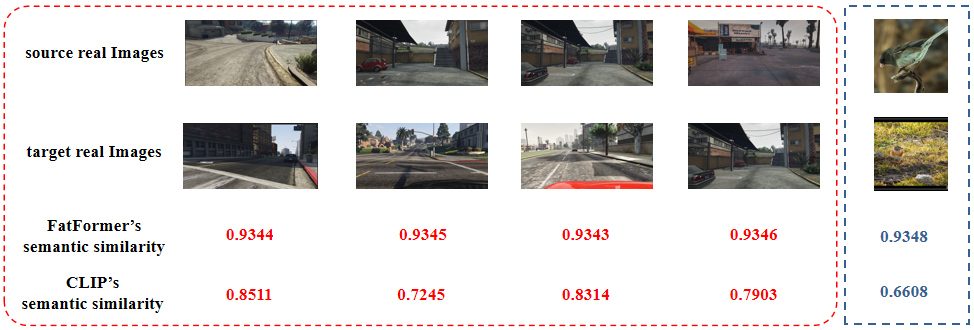}
    \caption{\textbf{Semantic similarity comparison of real images.} Inside the red dashed box, the source real images are correctly classified, while the target real images are misclassified. Inside the blue dashed box, both the source real images and the target real images are correctly classified. }
    \label{semantic2}
\end{figure*}



Previous works have validated the effectiveness of the CLIP model in the forgery detection task. Unlike conventional detection models, CLIP distinguishes itself by jointly learning from both visual and textual modalities, enabling it to understand and align images with natural language. This enables CLIP to better understand the semantic relationships between images and text, allowing for more nuanced detection of subtle forgery traces. In contrast, traditional detectors typically focus exclusively on the visual features of the images themselves, without leveraging additional semantic conceptual information. We attribute our preliminary findings to the influence of conceptual semantic factors, which help to distinguish real from fake images more effectively.

In Section 2, we briefly introduce the differences between the features extracted by CNNSpot \cite{CNNSpot} and CLIP \cite{UniFD}. In this section, we provide a more detailed discussion of these differences and investigate the characteristics of these differences and the reasons behind them. To further validate the distinguishing role of concepts in differentiating real and fake images, we observe the feature spaces of different categories of real and fake images. Both sets of features originate from the inputs of the detectors' final fully connected (FC) layers. This setup enables us to explore the distinction between concept-related features and those typically extracted by general detectors.

As illustrated in ~\cref{demograph}, we observe a notable difference in how real and fake images are represented in the visual semantic concept space. The gap between real and fake images in CLIP space is more pronounced across various categories, suggesting that semantic concepts help separate these two types of images more effectively. In contrast, in the feature space of CNNSpot, the distinction between real and fake images becomes much less obvious and more uniform, indicating that the learned features tend to exhibit monotonic patterns, which may lead to overfitting and limiting the model’s ability to generalize to unseen data. This highlights the importance of incorporating conceptual semantic understanding into the feature extraction process.

From these observations, we conclude that the use of concept-based features can significantly alleviate the problem of overfitting and improve a model's ability to generalize to unseen generative models.

\section{Analysis of Semantic Description Granularity in FatFormer}
\label{B}
In the introduction, we point out that the soft prompts based on simple [CLASS] embeddings of FatFormer have an intrinsic limitation in their semantic description granularity. This concern arises from our observation that FatFormer achieves a significantly lower $racc_m$ compared to UnivFD. The $racc_m$ is shown in ~\cref{racc}. This indicates that, when faced with real images, FatFormer is more likely to misclassify them as fake images. In more extreme terms, compared to its backbone, FatFormer appears to have lost the ability to recognize authentic images, which is clearly an anomalous behavior.

\begin{figure}[htbp]
    \centering
    \includegraphics[width=0.8\linewidth ]{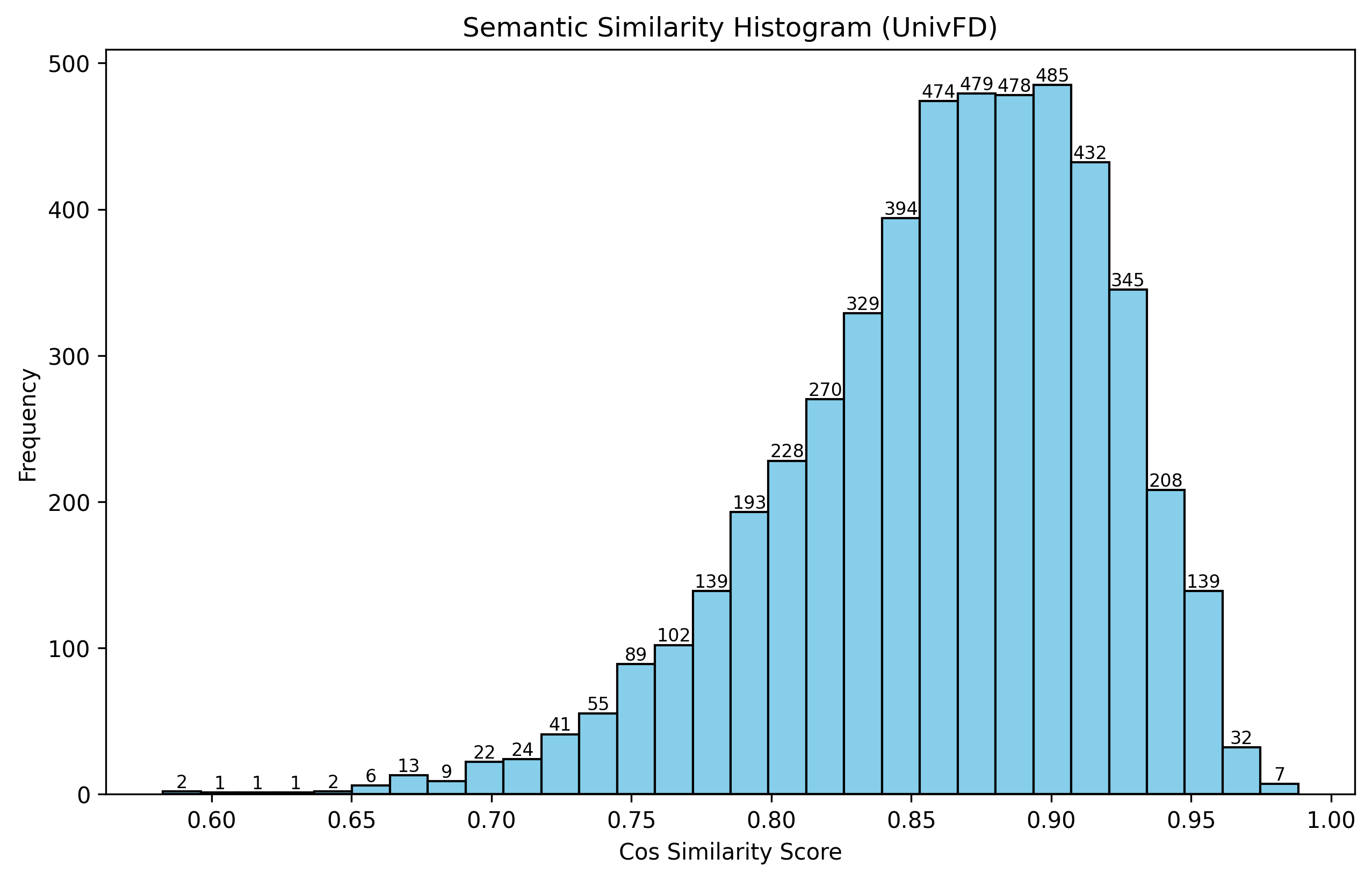}
    \caption{\textbf{Semantic similarity histogram of UnivFD.} The data is primarily concentrated in the cosine similarity range of 0.85 to 0.90, with the overall data falling within the range of 0.582 to 0.988. }
    \label{semantic}
\end{figure}

\begin{figure}[htbp]
    \centering
    \includegraphics[width=0.8\linewidth ]{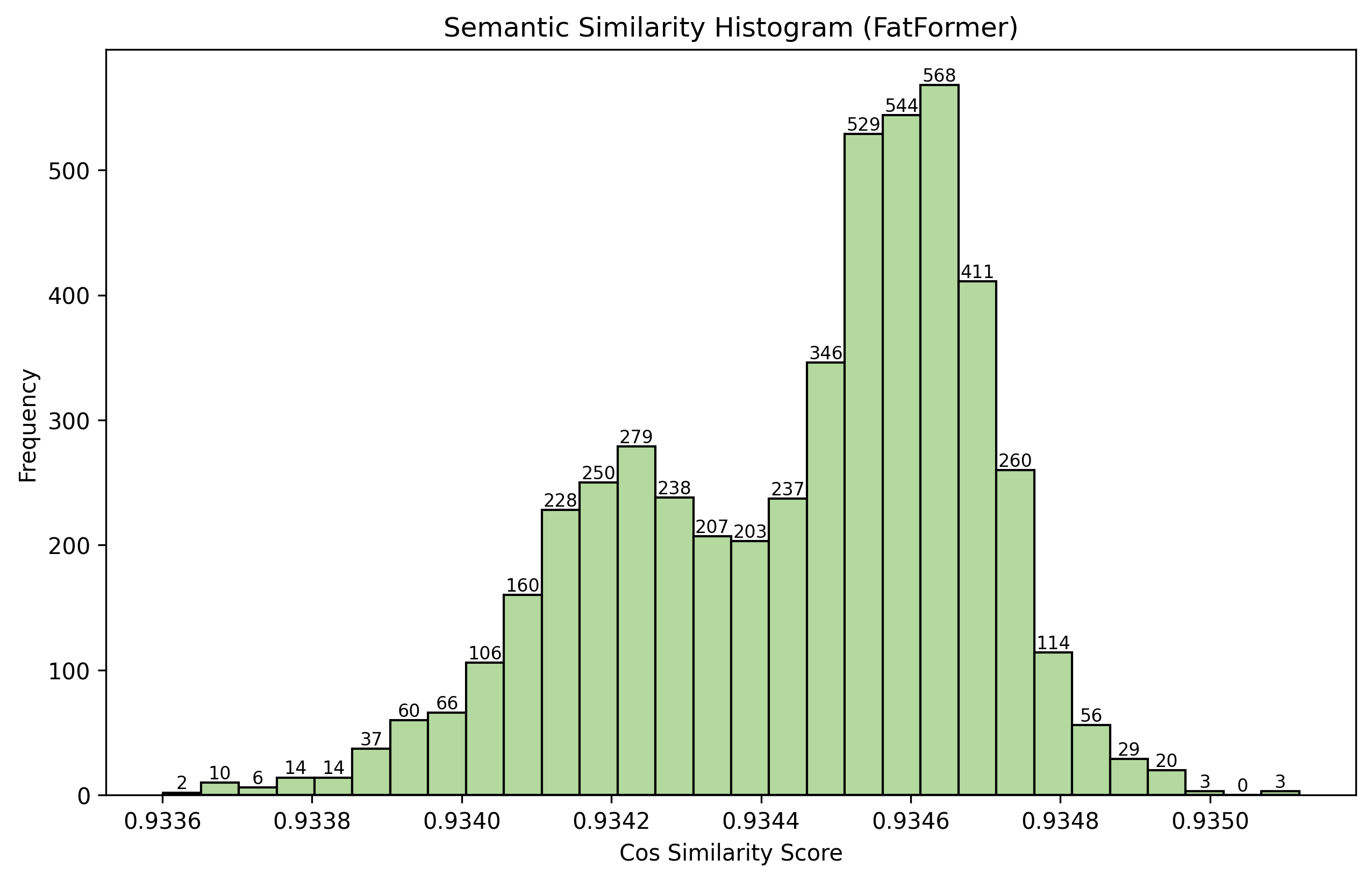}
    \caption{\textbf{Semantic similarity histogram of FatFormer.} The data is primarily concentrated in the cosine similarity range of 0.9344 to 0.9348, with the overall data falling within the range of 0.9336 to 0.9351.  }
    \label{semantic1}
\end{figure}

Our intuitive explanation is that the coarse-grained soft prompts used in FatFormer weaken its ability to perceive varying visual semantic details in real images. To validate this hypothesis, we randomly sampled 5,000 pairs of images and computed the cosine similarity between them based on the output vectors of FatFormer’s text encoder and the final-layer features of UniVFD’s image encoder. As shown in ~\cref{semantic1} and ~\cref{semantic}, the cosine similarity scores for UnivFD show a wider range of variation compared to FatFormer's. It indicates that FatFormer's soft prompts fail to distinguish semantic differences between images, indicating a significant decline in semantic discrimination capability.

Furthermore, we compared the semantic similarity between real images that were misclassified as fake and those that were correctly classified. As shown in ~\cref{semantic2}, despite the higher semantic similarity among real images, the UnivFD is still able to correctly determine their authenticity. In contrast, FatFormer, while eliminating semantic information interference, fails to make accurate authenticity judgments.

These findings suggest that although forgery-adaptive mechanisms improve FatFormer’s sensitivity to forgery traces, the lack of adequate semantic-guided information provided by the soft prompts hinders the model’s generalization ability in real-world scenarios.

\section{Training details}
In this section, we provide the details regarding the training process of our work. We use the official code repository provided by \cite{UniFD}. We train the CLIP:ViT variant of this baseline with Blur and JPEG augmentations applied with a probability of 0.5. The network is trained with a batch size of 32 and a learning rate of \(1 \times 10^{-4}\). The random seed is set to 46. For the loss function, the hyper-parameters \(\lambda_1\) and \(\lambda_2\) are set to $\frac{1}{9}$ and $\frac{1}{3}$, respectively. During testing, no Blur or JPEG augmentation is applied. Lastly, when training our classifier, we make use of Blur + JPEG data augmentations, any real or fake image is first augmented before being passed to the CLIP:ViT encoder (\(\varphi\)).

\section{Effect of sampling rate $\delta$ of SDL}
In this section, we investigate the effect of the parameter $\delta$ on forgery detection performance. We set $\delta$ to various values of \( \frac{1}{500}, \frac{1}{1000}, \frac{1}{2000}, \frac{1}{4000}, \frac{1}{5000} \) to explore how the number of sampled tokens impacts the detection task. Notably, our sampled dataset is drawn from the entire training set in \cite{UniFD}. Despite the large size of the training set, the number of sampled tokens remains below expectations — some segments contain no patch tokens at all.

\begin{figure}[tp]
    \centering
    \includegraphics[width=8cm]{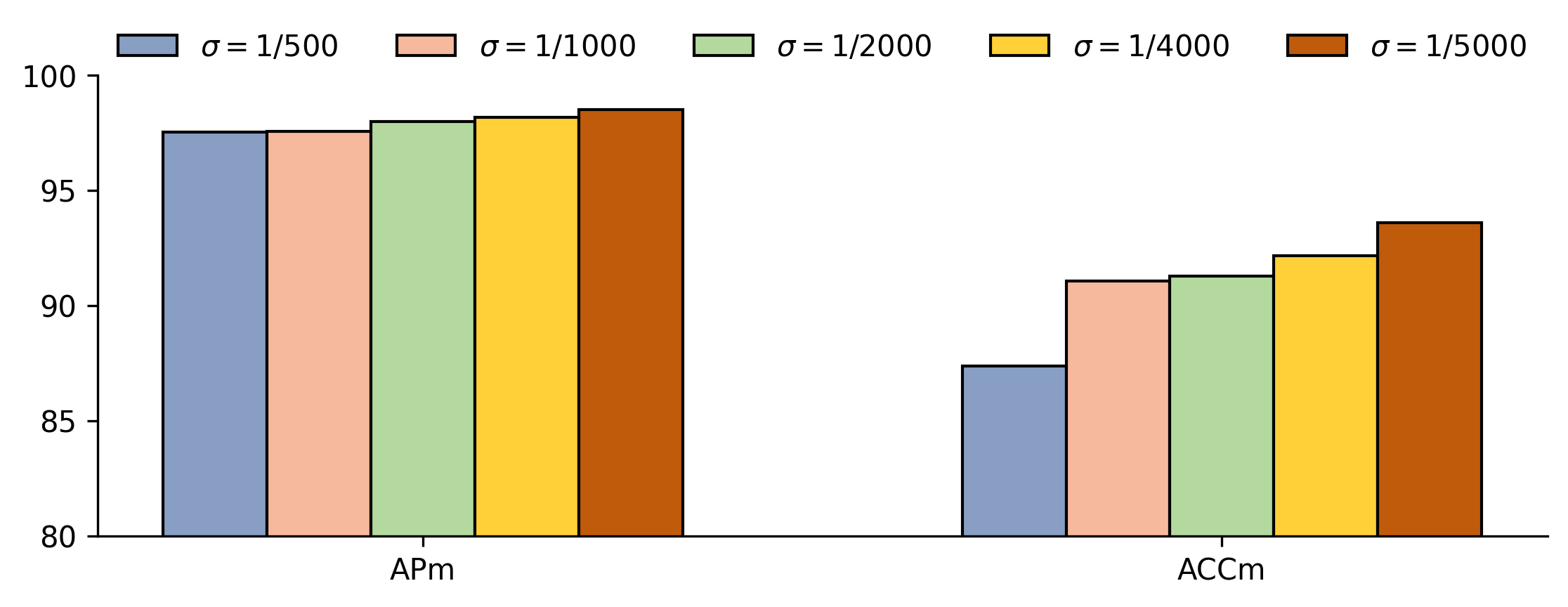}
    \caption{Performance of sampling rate $\delta$ of SDL.}
    \label{token}
\end{figure}

Intuitively, increasing the number of tokens should allow the model to better reflect the true distribution of visual semantic concepts, as more tokens provide a more comprehensive representation of the whole image. As shown in ~\cref{token}, when $\delta$ changes from \( \frac{1}{500} \) to \( \frac{1}{1000} \), although the change in Average Precision ($AP_m$) is not significantly large, there is a noticeable improvement in Accuracy ($ACC_m$), demonstrating that the additional tokens help the model better differentiate between real and fake images. 

Beyond this point, as $\delta$ continues to increase, the changes in both $AP_m$ and $ACC_m$ increase gradually, suggesting that after a certain threshold, increasing the number of sampled tokens yields diminishing returns in performance.  These findings underscore the effectiveness of the sampled tokens in enhancing the model's ability to detect forgery traces. 

Moreover, even with a relatively small number of tokens, the model achieves significant performance improvements. This characteristic is especially valuable as it reduces both computational costs and memory usage, making it a more efficient solution for real-world applications. In conclusion, this finding highlights that our method is both effective in detecting real and fake images and computationally efficient, even with fewer tokens.

\section{Robustness}
\begin{figure*}[htbp]
    \centering
    \includegraphics[width=\linewidth]{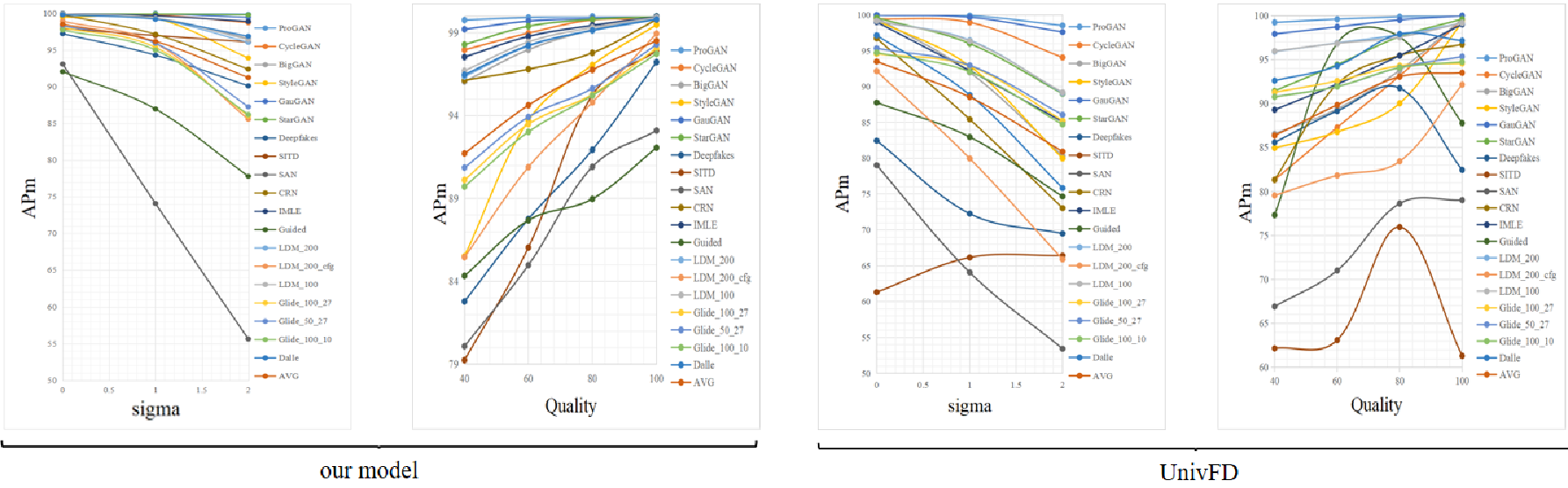}
    \caption{Robustness to different image processing operations. Both our detector and the trained baseline \cite{UniFD} demonstrate general robustness to these artifacts, but our performance is notably superior on unseen models. }
    \label{robustness}
\end{figure*}

In order to evade a fake detection system, an attacker may apply certain low-level post-processing operations to the fake images. To evaluate the robustness of our classifier against such operations, we follow prior work and assess its performance under different post-processing conditions. As shown in ~\cref{robustness}, our method demonstrates general robustness to both blur and JPEG compression artifacts compared to the baseline \cite{UniFD}.

It is worth noting that as the Gaussian blur sigma value changes, the average precision (AP) for different generative models consistently remains above $75\%$, with the exception of the SAN model. This indicates that our method is quite robust to Gaussian blur, effectively detecting forgery traces even under varying levels of blur. In particular, the AP remains stable across most generative models, suggesting that our approach maintains strong performance in the presence of noise or degradation typically introduced by Gaussian blur. However, for the SAN model, a noticeable drop in AP suggests that certain models, such as SAN, might be more sensitive to this type of distortion. 

On the other hand, when the JPEG compression quality is varied, the AP for all forgery models remains consistently above $80\%$, indicating that our method is highly resistant to JPEG compression artifacts. This is a strong indication of the model's ability to maintain accuracy even under common image compression techniques that often degrade the quality of forged images. Notably, our models exhibit minimal degradation in AP, which demonstrates their capability to accurately distinguish between real and fake images, even when compression artifacts are present. In contrast, models that are not robust to such distortions may experience significant drops in AP, reflecting their vulnerability to such post-processing operations.


\section{Accuracy breakdown of real and fake
classes}

Lastly, we break down the performance of different methods into performance on real (~\cref{racc}) and fake images (~\cref{facc}) associated with different generative models. This breakdown helps us understand the specific ways in which a detection method may fail. In particular, we observe that an image-level classifier, such as CNNSpot~\cite{CNNSpot}, works well in detecting real and fake images when they belong to the GAN domain. However, when tested on images from latent diffusion models, the network tends to classify almost all images as real. Consequently, while the classification accuracy on real images remains high, the accuracy on fake images drops drastically.

In contrast to other models, our method strikes a remarkable balance between performance on real and fake images, as evidenced by the results in ~\cref{facc} and ~\cref{racc}, where the fake image classification accuracy ($facc_m$) and real image classification accuracy ($racc_m$) are $93.16\%$ and $94.06\%$, respectively. This indicates that our model excels at distinguishing between real and fake images. Furthermore, our model has learned a feature space that effectively differentiates between these two categories. This ability to maintain consistent performance across both real and fake images highlights the robustness and effectiveness of our model in real-world applications. Our approach demonstrates its capability to detect subtle forgery traces, irrespective of the generative model used to create the fake images.

\begin{table*}[!t]
    \setlength{\tabcolsep}{2pt}
    \renewcommand{\arraystretch}{1.4}
    \centering
     \resizebox{\textwidth}{16mm}{
    \begin{tabular}{l l c c c c c c c c c c c c c c c c c c c c c}
        \hline
        \multirow{3}{*}{Methods} & \multirow{3}{*}{Ref} & \multicolumn{6}{c}{GAN} 
        & \multirow{3}{*}{\shortstack[c]{Deep \\ [1ex] fakes}}& \multicolumn{2}{c}{Low level}& \multicolumn{2}{c}{Perceptual loss}& \multirow{3}{*}{Guided}& \multicolumn{3}{c}{LDM}& \multicolumn{3}{c}{Glide}& \multirow{3}{*}{Dalle} & \multirow{3}{*}{Avg-acc} \\
        \cmidrule(r){3-8} \cmidrule(r){10-11} \cmidrule(r){12-13} \cmidrule(r){15-17} \cmidrule(r){18-20} 
        
         & & Pro- & Cycle-& Big-&Style- & Gau-& Star-&  &\multirow{2}{*}{SITD} & \multirow{2}{*}{SAN} &\multirow{2}{*}{CRN}  & \multirow{2}{*}{IMLE} &  & 200&200& 100& 100&50 & 100&  & \\

        & & GAN& GAN& GAN & GAN&GAN &GAN&  & &  &  &&  &Steps& w/cfg & Steps& 27 & 27& 10  & & \\
        
        \hline
        CNN-Spot&CVPR2020        &\textcolor{red}{100.0}&98.64&99.05&\underline{99.95}&99.40&99.30&\underline{99.45}&\textcolor{red}{100.0}&\textcolor{red}{100.0}&99.22&99.22&\underline{99.14}&\textcolor{red}{99.61}&\textcolor{red}{99.61}&\underline{99.61}&\textcolor{red}{99.61}&\textcolor{red}{99.61}&\textcolor{red}{99.61}&\textcolor{red}{99.61}&\underline{99.50}\\ 
        PatchFor&ECCV2020        &95.30&65.56&61.35&85.95&49.88&75.83&89.21&43.48&47.24&12.25&12.25&61.34&84.86&84.86&84.86&84.86&84.86&84.86&84.86&68.08\\ 
         
        Freq-spec&WIFS2019       &\underline{99.80}&\underline{99.80}&\underline{99.10}&99.90&\underline{99.80}&99.30&\textcolor{red}{100.0}&\textcolor{red}{100.0}&\textcolor{red}{100.0}&\underline{99.80}&\underline{99.80}&\textcolor{red}{99.60}&\underline{99.40}&\underline{99.40}&99.50&\underline{99.40}&\underline{99.50}&\underline{99.40}&\underline{99.60}&\textcolor{red}{99.60}\\ 
        
        UnivFD&CVPR2023&99.08&87.21&92.55&99.63&95.88&99.35&96.0&61.0&95.0&\underline{96.47}&96.47&93.34&92.39&92.39&92.39&92.39&92.39&92.39&92.39&92.56&\\
        FatFormer&CVPR2024&\textcolor{red}{100}&98.71&\underline{99.1}&\textcolor{red}{100}&98.88&\underline{99.50}&\underline{99.45}&63.3&\underline{98.17}&38.94&38.94&97.90&99.30&99.30&99.30&99.30&99.30&99.30&99.30&90.95&\\
        
       \textbf{SDD}&&\textcolor{red}{100.0}&\textcolor{red}{100.0}&\textcolor{red}{100.0}&\textcolor{red}{99.95}&\textcolor{red}{100.0}&\textcolor{red}{100.0}&{89.47}&\underline{99.44}&60.27&\textcolor{red}{99.86}&\textcolor{red}{100.0}&{65.50}&\underline{99.40}&{92.50}&\textcolor{red}{99.80}&87.70&90.0&86.90&{99.30}&{93.16}&\\
      
        \hline
       
    \end{tabular}}
    \caption{\textbf{Accuracy of detecting real images. } For each generative model (column), we consider its corresponding real images and test how
frequently a classifier (row) correctly predicts it as real.}
    \label{racc}
\end{table*}

\begin{table*}[!t]
    \setlength{\tabcolsep}{2pt}
    \renewcommand{\arraystretch}{1.4}
    \centering
     \resizebox{\textwidth}{16mm}{
    \begin{tabular}{l l c c c c c c c c c c c c c c c c c c c c c}
        \hline
        \multirow{3}{*}{Methods} & \multirow{3}{*}{Ref} & \multicolumn{6}{c}{GAN} 
        &\multirow{3}{*}{\shortstack[c]{Deep \\ [1ex] fakes}}& \multicolumn{2}{c}{Low level}& \multicolumn{2}{c}{Perceptual loss}& \multirow{3}{*}{Guided}& \multicolumn{3}{c}{LDM}& \multicolumn{3}{c}{Glide}& \multirow{3}{*}{Dalle} & \multirow{3}{*}{Avg-acc} \\
        \cmidrule(r){3-8} \cmidrule(r){10-11} \cmidrule(r){12-13} \cmidrule(r){15-17} \cmidrule(r){18-20} 
        
         & & Pro- & Cycle-& Big-&Style- & Gau-& Star-&  &\multirow{2}{*}{SITD} & \multirow{2}{*}{SAN} &\multirow{2}{*}{CRN}  & \multirow{2}{*}{IMLE} &  & 200&200& 100& 100&50 & 100&  & \\

        & & GAN& GAN& GAN & GAN&GAN &GAN&  & &  &  &&  &Steps& w/cfg & Steps& 27 & 27& 10  & & \\
        
        \hline
        CNN-Spot&CVPR2020        &\textcolor{red}{100.0}&62.91&18.90&38.52&59.10&62.58&2.52&13.89&0.0&75.95&88.92&4.67&3.05&4.26&2.96&9.25&12.34&9.1&4.9&30.20\\ 
        PatchFor&ECCV2020        &93.45&69.20&67.90&78.56&64.51&84.74&21.31&\underline{85.70}&\underline{54.90}&96.33&97.96&\underline{68.94}&73.32&\underline{67.48}&73.86&49.26&52.23&51.22&54.02&68.68\\ 
         
        Freq-spec&WIFS2019       &0.20&\textcolor{red}{100.0}&1.80&0.0&0.90&\textcolor{red}{100.0}&0.0&0.0&0.4&1.30&0.50&0.40&1.30&1.40&1.10&3.90&3.30&1.30&0.50&11.50\\ 
        
        UnivFD&CVPR2023&\textcolor{red}{100.0}&\underline{99.77}&84.7&61.88&\underline{98.34}&\underline{98.6}&73.0&82.0&27.0&42.06&61.94&48.77&\underline{90.2}&51.65&90.2&85.7&88.94&87.77&70.55&75.95&\\
        FatFormer&CVPR2024&99.78&\textcolor{red}{100}&\textcolor{red}{99.90}&\textcolor{red}{94.24}&\textcolor{red}{99.98}&\textcolor{red}{100}&\underline{87.83}&\textcolor{red}{99.44}&37.90&\textcolor{red}{100}&\textcolor{red}{100}&54.10&\textcolor{red}{97.80}&\textcolor{red}{97.90}&\underline{90.40}&\underline{89.30}&\underline{89.90}&\underline{89.00}&\textcolor{red}{98.10}&\underline{90.78}&\\
        \textbf{SDD}&&\underline{99.75}&91.52&93.40&\underline{90.59}&96.92&98.35&\textcolor{red}{96.16}&67.78&\textcolor{red}{96.35}&92.95&92.95&\textcolor{red}{93.60}&\textcolor{red}{96.70}&\textcolor{red}{96.70}&\textcolor{red}{96.70}&\textcolor{red}{96.70}&\textcolor{red}{96.70}&\textcolor{red}{96.70}&\underline{96.70}&\textcolor{red}{94.06}\\
        \hline
    \end{tabular}}
    \caption{\textbf{Accuracy of detecting fake images.} For each generative model (column), we consider its corresponding fake images and test
how frequently a classifier (row) correctly predicts it as fake.}
    \label{facc}
\end{table*}

\begin{figure*}[htbp]
    \centering
    \includegraphics[width=\linewidth ]{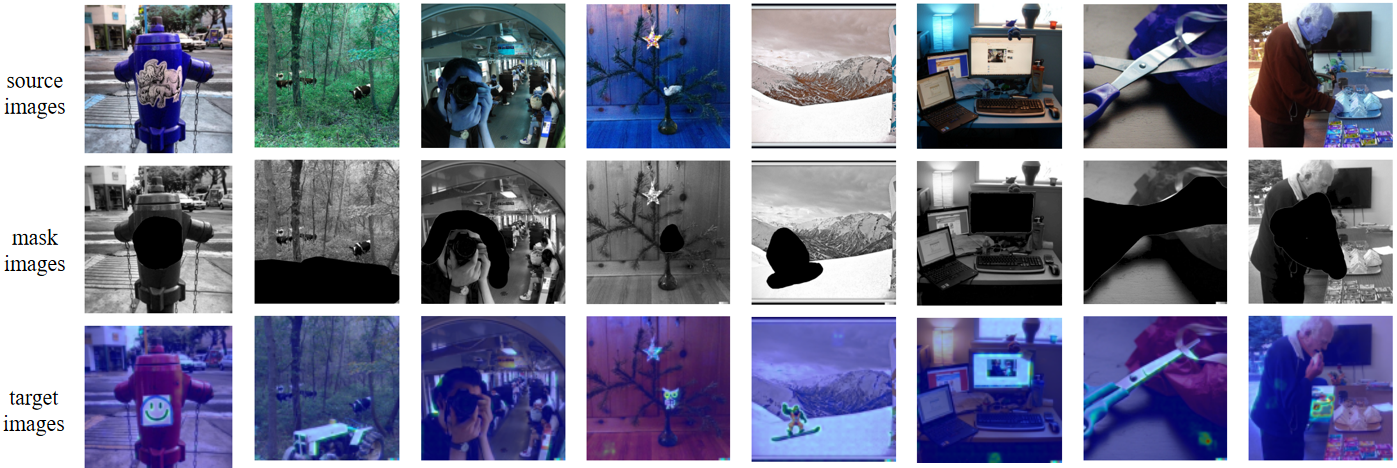}
    \caption{ The visualization of attention on dataset \cite{PerceptualAL}.
    The first row displays the original image, the second row shows the corresponding mask, and the third row presents the generated image within the masked region.}
    \label{demograph}
\end{figure*}

\begin{table}
    \centering
  \begin{tabular}{ccc}
    \toprule
    {Method} & $AP_{m}$ & $Acc_{m}$ \\ 
    \midrule
    BLIP & \multirow{2}{*}{95.61} & \multirow{2}{*}{84
    .46} \\
    (VIT-L/16)&&    \\ 
    CLIP & \multirow{2}{*}{98.52} & \multirow{2}{*}{93.61} \\
    (VIT-L/14)&&   \\
    \bottomrule
    \end{tabular}
    \captionof{table}{Comparisons with different backbones on the UnivFD dataset. }
    \label{blip}
\end{table}

\section{ Comparisons with different backbones on the UnivFD dataset.}
To further investigate the role of semantic concepts, we adopt the BLIP: VIT-L/16 as the backbone for forgery detection. We hypothesize that BLIP provides stronger fine-grained perception over the entire image, potentially making it more suitable for capturing semantic-level inconsistencies in manipulated content. Unlike CLIP, which primarily focuses on contrastive learning, BLIP is trained using vision-language pretraining tasks such as image-text matching and image captioning, leading to improved vision-language alignment and a more detailed semantic understanding. During the experiment, we observed that the number of patch tokens sampled by BLIP is fewer than that by CLIP. This seems to suggest the incompleteness and inadequacy of BLIP's visual semantic concept space.

However, as shown in ~\cref{blip}, using CLIP as the backbone yields better performance than using BLIP, which deepened our understanding of the semantic concept space. Despite its stronger alignment at the image-caption level, BLIP appears to have a less comprehensive and diverse concept space compared to CLIP, resulting in concept-forgery misalignment.

We attribute this limitation primarily to the scale and diversity of pretraining data. BLIP is trained on 129M samples, while CLIP uses 400M samples. The broader and more diverse supervision in CLIP likely equips it with a more robust and generalizable semantic embedding space, especially under open-world or adversarial conditions such as image forgery. Furthermore, CLIP’s contrastive training may emphasize discriminative concept boundaries, which could be inherently more beneficial for tasks requiring semantic-level anomaly detection.

In summary, although BLIP possesses advantages in fine-grained alignment and descriptive representation, its current pretraining scale and objectives may limit its effectiveness in tasks like forgery detection, where broad semantic coverage and discriminative representation are critical.

\section{Effect of SDD on the tampered dataset\cite{tamper}}
Beyond simply classifying images as real or generated, numerous research efforts have sought to localize the edited regions within the tampered images. Since we emphasize the role of CFDL in localizing semantically relevant forgery regions in this work, we try to apply our pretrained model trained on Stable Diffusion v1 images and random real LAION images to detect manipulated regions in tampered images. Clearly, identifying the authenticity of a whole image becomes a significant challenge due to the increasing proportion of real content within a given image. To further investigate our model's ability to tackle this challenge, we conduct experiments on the MAGICBRUSH dataset \cite{tamper}. MAGICBRUSH, finetuned by InstructPix2Pix, is a manually annotated dataset for instruction-guided real image editing that covers diverse scenarios: single-turn, multi-turn, mask-provided, and mask-free editing. 

We input tampered images into our model and obtain the corresponding heatmaps using CAM. Although our model's forgery detection performance decreases on this dataset, by analyzing the heatmaps alongside the mask images, we are surprised to find that our model can still localize the manipulated regions, albeit with limited accuracy. This demonstrates the significant potential of our model in the field of image forgery detection. In the future, we plan to further explore methods for distinguishing fake images that have been manipulated from real images.
\begin{table}
\centering
\begin{tabular}{cccc}
    \toprule
    \multirow{2}{*}{Model} & \multirow{2}{*}{FPS} & \multirow{2}{*}{\vspace*{0.3cm}Time} & \multirow{2}{*}{\textbf{\vspace*{0.3cm}GPU us- }} \\
    & & (ms) & age(MB) \\ 
    \midrule
    SDD  &15&68&3555\\ 
    -LORA &17&59&3252\\ 
    -feature enhancement &16&61&3186\\ 
    -LA &16&63&3510\\
    \bottomrule
    \end{tabular}
    \vspace{-0.3cm}
    \captionof{table}{The computational cost of our model without different modules on the UnivFD dataset. The prefix ‘-’ indicates the module is removed.} 
    \label{cost}
\end{table}

\section{ More analysis of learned latent space}

\begin{figure}[htbp]
    \centering
    \includegraphics[width=\linewidth]{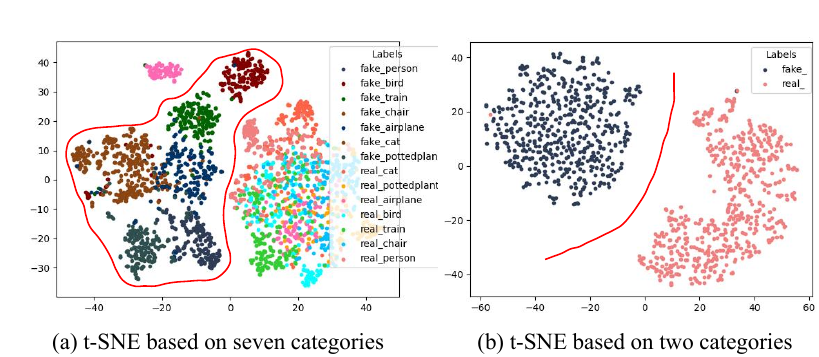}
    \vspace{-0.6cm}
    \caption{The t-SNE visualization of semantic concepts with different numbers of categories, where the size of samples is equal.}
    \label{tsne_O}
\end{figure}

We argue that the indistinct boundaries observed in generative models arise from applying t-SNE across multiple classes (i.e., semantic concepts), while the visualization itself is presented in a binary fashion (real vs. fake). To further support this claim, we perform a more fine-grained t-SNE analysis on the ProGAN test data (only the ProGAN dataset provides explicit class labels for each sample. Other generative models do not offer such semantic annotations) using explicit class labels. In particular, we visualize the feature distribution of samples from a combined subset of categories — person, bird, train, chair, airplane, cat, and potted plant — as well ajhjhs from the airplane category alone.

As shown in ~\cref{tsne_O}, increasing concept diversity leads to blurrier global boundaries in the t-SNE projection. Nevertheless, real and fake samples within the same concept remain locally separable, suggesting that the observed structure is shaped by concept-aware organization.

\section{ The computational cost of our modules}
We evaluate the computational cost introduced by our key components: LoRA fine-tuning (LORA), feature enhancement, and reconstruction-based alignment (RA). As shown in ~\cref{cost}, all three modules introduce only a minor increase in inference-time cost, maintaining the model's efficiency while improving performance.

Specifically, the full model (SDD) runs at 15 FPS with an average inference time of 68 ms and a GPU memory footprint of 3555 MB. Removing LoRA slightly improves FPS to 17 and reduces memory usage by approximately 300 MB, indicating that LoRA contributes a small computational cost. Removing feature enhancement results in the lowest memory usage (3186 MB) and a slight FPS increase, showing that multi-scale feature fusion is lightweight in practice. Excluding the RA module also reduces the inference time slightly, suggesting that reconstruction-based alignment introduces minimal cost while contributing important semantic consistency.

Overall, these results confirm that our proposed components are computationally efficient and practical for real-world deployment scenarios, striking a favorable balance between performance and resource consumption.
\newpage